\documentclass{article}

\usepackage[english]{babel}

\usepackage[letterpaper,top=2cm,bottom=2cm,left=3cm,right=3cm,marginparwidth=1.75cm]{geometry}

\usepackage{microtype}
\usepackage{graphicx}
\usepackage{subcaption}
\usepackage{booktabs} 
\usepackage{tabularx}

\usepackage{amsmath}
\usepackage{amssymb}
\usepackage{mathtools}
\usepackage{amsthm}
\usepackage[colorlinks=true, allcolors=blue]{hyperref}

\usepackage{algorithm,algorithmic}

\theoremstyle{plain}
\newtheorem{theorem}{Theorem}[section]
\newtheorem{proposition}[theorem]{Proposition}

\theoremstyle{definition}

\theoremstyle{remark}

\allowdisplaybreaks

\title{LSB: Local Self-Balancing MCMC in Discrete Spaces}
\author{Emanuele Sansone\thanks{emanuele.sansone@kuleuven.be}\\
Department of Computer Science\\
KU Leuven}
\date{}

\begin{document}
\maketitle

\begin{abstract}
We present the Local Self-Balancing sampler (LSB), a local Markov Chain Monte Carlo (MCMC) method for sampling in purely discrete domains, which is able to autonomously adapt to the target distribution and to reduce the number of target evaluations required to converge. LSB is based on (i) a parametrization of locally balanced proposals, (ii) a newly proposed objective function based on mutual information and (iii) a self-balancing learning procedure, which minimises the proposed objective to update the proposal parameters. Experiments on energy-based models and Markov networks show that LSB converges using a smaller number of queries to the oracle distribution compared to recent local MCMC samplers.
\end{abstract}


{\section{Introduction}
	\label{sec:intro}}
Sampling from complex and intractable probability distributions is of fundamental importance for learning and inference~\cite{mackay2003information}. In this regard, MCMC methods are promising solutions to tackle the intractability of sampling in high dimensions. They have been successfully applied in several domains, including Bayesian statistics and statistical physics~\cite{neal1993probabilistic,robert2013monte},  bioinformatics and computational biology~\cite{altekar2004parallel,alexeev2020markov} as well as machine learning~\cite{andrieu2003introduction,koller2009probabilistic,nijkamp2020anatomy}.

While MCMC is a general method, which can be used to sample from any target distribution, its efficiency strongly depends on the choice of the proposal. Indeed, the proposal must be adapted to the distribution we want to sample from~\cite{andrieu2008tutorial,hoffman2014no} and machine learning can help to automate this process~\cite{zhang2012continuous,pakman2013auxiliary,afshar2015reflection,dinh2017probabilistic,nishimura2020discontinuous}. However, less effort has been devoted to devise strategies for adapting the proposal distribution to a discrete target. Most common solutions embed the discrete problem into a continuous one, thus allowing to reuse sampling strategies designed for the continuous domain. However, these strategies are not always optimal, as they either require specific analytic forms for the target ~\cite{zhang2012continuous}, or simply because the embeddings do not capture the topological properties of the original domain~\cite{pakman2013auxiliary,afshar2015reflection,dinh2017probabilistic,nishimura2020discontinuous}. We can avoid these issues by sampling directly in the discrete domain and using local MCMC strategies~\cite{zanella2020informed}.

In this work, we propose an adaptation strategy for local MCMC. Specifically, we introduce (i) a new parametrization for locally balanced proposals, (ii) a new objective function based on mutual information to compute the statistical dependence between consecutive samples, and (iii) a learning procedure to adapt the parameters of the proposal to the target distribution using our proposed objective. The resulting procedure, called the Local Self-Balancing sampler (LSB), automatically discovers a locally balanced proposal with the advantage of reducing the number of target evaluations required to converge. Furthermore, we show that our framework generalizes over two recent works~\cite{zanella2020informed} and~\cite{grathwohl2021oops}. We conduct experiments on energy-based models and Markov networks, and show that LSB queries the oracle distribution in a more efficient way compared to recent local MCMC samplers.

In summary, the key contributions of the work are:
\begin{itemize}
    \item A parametrization of locally balanced proposals, which enables to reduce proposal selection into a learning problem.
    \item The use of mutual information as a criterion to learn the proposal distribution and accelerate MCMC.
    \item An estimator to efficiently compute the mutual information objective and to enable efficient learning.
\end{itemize}

We start by providing some background on locally balanced proposal distributions (Section~\ref{sec:back}), we introduce LSB by describing the parametrizations, the objective and the learning procedure (Section~\ref{sec:lsb}), we discuss the related work (Section~\ref{sec:related}) and the experiments (Section~\ref{sec:experiments}), and finally we conclude by highlighting the main limitations of LSB and the possible future directions (Section~\ref{sec:conclusion}).

{\section{Background}
	\label{sec:back}}
We consider the problem of sampling from a distribution $p$ with a support defined over a large finite discrete sample space $\mathcal{X}$, i.e. $p(\mathbf{x})=\tilde{p}(\mathbf{x})/\sum_{\mathbf{x}''\in\mathcal{X}}\tilde{p}(\mathbf{x}'')$, where the normalization term cannot be tractably computed and only $\tilde{p}$ can be evaluated. One solution to the problem consists of sampling using MCMC~\cite{neal1993probabilistic}. The main idea of MCMC is to sequentially sample from a tractable surrogate distribution, alternatively called proposal, and to use an acceptance criterion to ensure that generated samples are distributed according to the original distribution. More formally, MCMC is a Markov chain with a transition probability of the form:\footnote{For all $\mathbf{x},\mathbf{x}'\in\mathcal{X}$ such that $\mathbf{x}\neq\mathbf{x}'$, see the Supplementary material for the complete form}
\begin{align}
    T(\mathbf{x'}|\mathbf{x}) =A(\mathbf{x'},\mathbf{x})Q(\mathbf{x'}|\mathbf{x})
    \label{eq:transition}
\end{align}
where $Q(\mathbf{x'}|\mathbf{x})$ is the probability of sampling $\mathbf{x'}$ given a previously sampled $\mathbf{x}$, namely the proposal distribution, and $A(\mathbf{x'},\mathbf{x})$ is the probability of accepting sample $\mathbf{x'}$ given $\mathbf{x}$, e.g. $A(\mathbf{x'},\mathbf{x})=\min\big\{1,\frac{\tilde{p}(\mathbf{x'})Q(\mathbf{x}|\mathbf{x'})}{\tilde{p}(\mathbf{x})Q(\mathbf{x'}|\mathbf{x})}\big\}$.\footnote{Other choices are available~\cite{neal1993probabilistic} as well.}
In this work, we consider the family of locally informed proposals~\cite{zanella2020informed}, which are characterized by the following expression:
\begin{equation}
    Q(\mathbf{x'}|\mathbf{x}) = \frac{g\big(\frac{\tilde{p}(\mathbf{x'})}{\tilde{p}(\mathbf{x})}\big)1[\mathbf{x'}\in N(\mathbf{x})]}{Z(\mathbf{x})}
    \label{eq:proposal}
\end{equation}
where $N(\mathbf{x})$ is the neighborhood of $\mathbf{x}$ based on the Hamming metric.\footnote{In this work, $N(\mathbf{x})$ is a set of points having Hamming distance 1 from $\mathbf{x}$.}

Note that the choice of $g$ has a dramatic impact on the performance of the Markov chain, as investigated in~\cite{zanella2020informed}. In fact, there is a family of functions called \emph{balancing functions}, satisfying the relation $g(t)=tg(1/t)$ (for all $t>0$), which have extremely desirable properties, briefly recalled hereunder.

\paragraph{Acceptance rate.} The balancing property allows to rewrite the acceptance function in a form that depends only from the ratio $\frac{Z(\mathbf{x})}{Z(\mathbf{x'})}$, specifically 
$A(\mathbf{x'},\mathbf{x})=\min\big\{1,\frac{Z(\mathbf{x})}{Z(\mathbf{x'})}\big\}$. Therefore if $\tilde{p}(\mathbf{x})$ is smooth enough, then the ratio $\frac{Z(\mathbf{x})}{Z(\mathbf{x'})}$ will be close to 1, as $\mathbf{x},\mathbf{x'}$ are close to each other. However, a proper choice of $g$ can additionally influence the value of this ratio.

\paragraph{Detailed balance.} Note that for all $\mathbf{x'}=\mathbf{x}$, detailed balance trivially holds, viz. $p(\mathbf{x})T(\mathbf{x'}|\mathbf{x})=p(\mathbf{x'})T(\mathbf{x}|\mathbf{x'})$. In all other cases, detailed balance can be proved, by exploiting the fact that $T(\mathbf{x'}|\mathbf{x})=A(\mathbf{x'},\mathbf{x})Q(\mathbf{x'}|\mathbf{x})$ and by using the balancing property (see the Supplementary material for more details). Detailed balance is a sufficient condition for invariance. Consequently, the target $p$ is a fixed point of the Markov chain.

\paragraph{Ergodicity.} Under mild assumptions, we have also ergodicity (we leave more detailed discussion to the Supplementary material). In other words, the Markov chain converges to the fixed point $p$ independently from its initialization.

\paragraph{Efficiency.} The efficiency of MCMC is generally measured in terms of the resulting asymptotic variance for sample mean estimators. This is indeed a proxy to quantify the level of correlation between samples generated through MCMC. Higher levels of asymptotic variance correspond to higher levels of correlation, meaning that the Markov chain produces more dependent samples and it is therefore less efficient. Balancing functions are asymptotically optimal according to Peskun ordering~\cite{zanella2020informed}.

The work in~\cite{zanella2020informed} proposes a pool of balancing functions with closed-form expression together with some general guidelines to choose one. However, this pool is only a subset of the whole family of balancing functions and several cases do not even have an analytical expression. Consequently, it is not clear which function to use in order to sample efficiently from the target distribution.
Indeed, we will see in the experimental section that (i) the optimality of the balancing function depends on the target distribution and that (ii) in some cases the optimal balancing function may be different from the ones proposed in~\cite{zanella2020informed}. In the next sections, we propose a strategy to automatically learn the balancing function from the target distribution, thus reducing the number of target evaluations in order to achieve convergence compared to recent discrete MCMC samplers.

{\section{LSB: Local Self-Balancing Strategy}
	\label{sec:lsb}}
We start by introducing two different parametrizations for the family of balancing functions in increasing order of functional expressiveness. Then, we propose an objective criterion based on mutual information to learn the parametrization and to reduce the number of steps required to converge. Finally, we introduce an approximate strategy to further reduce the computational overhead of each step.

\subsection{Parametrizations}
We state the following proposition and then use it to devise the first parametrization.
\begin{proposition}
Given $n$ balancing functions $\mathbf{g}(t)=[g_1(t),\dots,g_n(t)]^T$ and a vector of scalar positive weights $\mathbf{w}=[w_1,\dots,x_n]^T$, the linear combination $g(t)\doteq\mathbf{w}^T\mathbf{g}(t)$ satisfies the balancing property.
\end{proposition}
\begin{proof}
    $g(t)=\mathbf{w}^T\mathbf{g}(t)=\sum_{i=1}^n w_i g_i(t)=t\sum_{i=1}^n w_i g_i(1/t)=t\mathbf{w}^T\mathbf{g}(1/t)=tg(1/t)$
\end{proof}
Despite its simplicity, the proposition has important implications. First of all, it allows to convert the problem of choosing the optimal balancing function into a learning problem. Secondly, the linear combination introduces only few parameters (in the experiments we consider $n=4$) and therefore the learning problem can be solved in an efficient way. The requirement about positive weights is necessary to guarantee ergodicity (see Supplementary material on ergodicity for further details).

The first parametrization (LSB 1) consists of the relations $w_i=e^{\theta_i}/\sum_{j=1}^n e^{\theta_j}$ for all $i=1,\dots,n$, where $\pmb{\theta}=[\theta_1,\dots,\theta_n]\in\mathbb{R}^n$. Note that the softmax is used to smoothly select one among the $n$ balancing functions. Therefore, we refer to this parametrization as learning to select among existing balancing functions.

The second parametrization (LSB 2) is obtained from the following proposition.
\begin{proposition}
Given $g_{\pmb{\theta}}(t)=\frac{h_{\pmb{\theta}}(t)}{2}+\frac{th_{\pmb{\theta}}(1/t)}{2}$, where $h_{\pmb{\theta}}$ is a universal real valued function approximator parameterized by vector $\pmb{\theta}\in\mathbb{R}^k$ (e.g. a neural network), and any balancing function $\ell$, there always exists $\tilde{\pmb{\theta}}\in\mathbb{R}^k$ such that $g_{\tilde{\pmb{\theta}}}(t)=\ell(t)$ for all $t>0$.
\end{proposition}
\begin{proof}
Given any balancing function $\ell$, we can always find a $\tilde{\pmb{\theta}}$ such that $h_{\tilde{\pmb{\theta}}}(t)=\ell(t)$ for all $t>0$ (because $h_{\pmb{\theta}}$ is a universal function approximator). This implies that $h_{\tilde{\pmb{\theta}}}$ satisfies the balancing property, i.e. $h_{\tilde{\pmb{\theta}}}(t)=t h_{\tilde{\pmb{\theta}}}(1/t)$ for all $t>0$. Consequently, by definition of $g_{\pmb{\theta}}$, we have that $g_{\pmb{\theta}}(t)=h_{\tilde{\pmb{\theta}}}(t)$. And finally we can conclude that $g_{\pmb{\theta}}(t)=\ell(t)$ for all $t>0$.
\end{proof}
The proposition enables to parameterize the whole family of balancing functions, thus generalizing the result obtained for LSB 1. Note that the increased level of expressiveness of LSB 2 comes at the cost of a higher computation. However, in practice we can always trade-off expressiveness and computation by specifying the capacity of the function approximator (e.g. the hyperparameters of a neural network). We leave this discussion to the experimental section.

In the next paragraphs, we propose an objective and a learning strategy to train the parameters of LSB 1 and LSB 2.

\subsection{Objective and Learning Algorithm}
The goal here is to devise a criterion to find the balancing function reducing the number of target likelihood evaluations and increasing the convergence/mixing rate to distribution $p$. 

As already mentioned in previous section, MCMC based on locally informed proposal converges to the true target distribution independently of its initialization. Furthermore, the rate of convergence and mixing is controlled by the balancing function. Importantly, we can speedup convergence and mixing by minimizing the amount of statistical dependence between consecutive samples. To this purpose, we introduce the following mutual information-based criterion:
%
\begin{align}
    \mathcal{I}_{\pmb{\theta}}=KL\big\{p(\pmb{x})T_{\pmb{\theta}}(\pmb{x}'|\pmb{x})\|p(\pmb{x})p(\pmb{x}')\big\}
    \label{eq:mutual}
\end{align}
where $KL$ is the Kullback Leibler divergence and $T_{\pmb{\theta}}$ is the transition probability with explicit dependence on parameter $\pmb{\theta}$. Now, we are ready to highlight some properties of Eq.~\ref{eq:mutual} with the following theorem (see Supplementary material for its proof).
\begin{theorem}
Consider $\mathcal{X}=\{0,1\}^d$ and $p(\pmb{x})=\tilde{p}(\pmb{x})/\Gamma$, where $\Gamma$ is a normalizing constant. Define
two auxiliary distributions $Q_1$ and $Q_2$, such that for all $\pmb{x}\in\mathcal{X}$, $Q_1(\pmb{x})>0$ whenever $\tilde{p}(\pmb{x})>0$, and for all $\pmb{x}'\in\mathcal{N(\pmb{x})}$, $Q_2(\pmb{x}')>0$ whenever $Q(\pmb{x}'|\pmb{x})>0$.
Therefore,
\begin{itemize}
    \item[(a)] if $\mathcal{M}(\pmb{x})\doteq 1-\sum_{\pmb{x}''\in N(\pmb{x})}A(\pmb{x}'',\pmb{x})Q(\pmb{x}''|\pmb{x})$, then
    \begin{align}
        \mathcal{L_{\pmb{\theta}}} =& E_{\substack{\pmb{x}\sim Q_1 \\ \pmb{x}'\sim Q_2}}\bigg\{\frac{\tilde{p}(\pmb{x})Q(\pmb{x}'|\pmb{x})}{Q_1(\pmb{x})Q_2(\pmb{x}')}A(\pmb{x}',\pmb{x})\log\frac{A(\pmb{x}',\pmb{x})Q(\pmb{x}'|\pmb{x})}{\tilde{p}(\pmb{x}')}\bigg\}+ E_{\pmb{x}\sim Q_1}\bigg\{\frac{\tilde{p}(\pmb{x})}{Q_1(\pmb{x})}\mathcal{M}(\pmb{x})\log\frac{\mathcal{M}(\pmb{x})}{\tilde{p}(\pmb{x})}\bigg\}
    \end{align}
    and $\mathcal{I}_{\pmb{\theta}}=\frac{\mathcal{L_{\pmb{\theta}}}}{\Gamma} + \frac{\log\Gamma}{\Gamma}$.
    \item[(b)] if $\mathcal{M}(\pmb{x})\doteq 1-A(\pmb{x}^*,\pmb{x})Q(\pmb{x}^*|\pmb{x})$, where $\pmb{x}^*$ is randomly sampled according to a uniform distribution over $N(\pmb{x})$, then
    \begin{align}
        \mathcal{L_{\pmb{\theta}}} =& E_{\substack{\pmb{x}\sim Q_1 \\ \pmb{x}'\sim Q_2}}\bigg\{\frac{\tilde{p}(\pmb{x})Q(\pmb{x}'|\pmb{x})}{Q_1(\pmb{x})Q_2(\pmb{x}')}A(\pmb{x}',\pmb{x})\log\frac{A(\pmb{x}',\pmb{x})Q(\pmb{x}'|\pmb{x})}{\tilde{p}(\pmb{x}')}\bigg\} + E_{\pmb{x}\sim Q_1}\bigg\{\frac{\mathcal{M}(\pmb{x})}{Q_1(\pmb{x})}\big[\eta \mathcal{M}(\pmb{x}) - \tilde{p}(\pmb{x})(\log\eta  +1)\big]\bigg\}
    \end{align}
    and $\mathcal{I}_{\pmb{\theta}}\leq\frac{\mathcal{L_{\pmb{\theta}}}}{\Gamma} + \frac{\log\Gamma}{\Gamma}$, for $\eta\in\mathbb{R}^+$.
\end{itemize}
\end{theorem}
\begin{algorithm}[tb]
   \caption{Local Self-Balancing (LSB)}
   \label{alg:lsb}
\begin{algorithmic}
   \STATE {\bfseries Input:} Learning rate $\gamma=1e-2$, $\pi=1e-8$, initial parameter $\pmb{\theta}_0$, burn-in iterations $K$ and batch of samples $N$ 
   \STATE $\{\hat{\pmb{x}}^{(i)}\}_{i=1}^N\sim U_{\mathcal{X}}$
   \FOR{$k=1$ {\bfseries to} $K$}
    \STATE $\{\pmb{x}^{(i)}\}_{i=1}^N\sim Q_1$
    \STATE $\{\pmb{x}'^{(i)}\}_{i=1}^N\sim Q_2$
    \STATE $\widehat{\mathcal{L}}_{\pmb{\theta}}\leftarrow$ Estimate $\mathcal{L}_{\pmb{\theta}}$ using $\{\pmb{x}^{(i)}\}_{i=1}^N$, $\{\pmb{x}'^{(i)}\}_{i=1}^N$
    \STATE $\pmb{\theta} \leftarrow \pmb{\theta}-\frac{\gamma}{N}\nabla_{\pmb{\theta}}\widehat{\mathcal{L}}_{\pmb{\theta}}$
    \STATE Update $\eta$
    \STATE Accept/Reject samples
    \STATE Update $\{\hat{\pmb{x}}^{(i)}\}_{i=1}^N$ with accepted samples
    \STATE $\pmb{\theta}_0\leftarrow\pmb{\theta}$
   \ENDFOR
\end{algorithmic}
\end{algorithm}

The theorem tells that for case (a), $\mathcal{L}_{\pmb{\theta}}$ is equal to $\mathcal{I}_{\pmb{\theta}}$ up to a constant, whereas, for case (b) $\mathcal{L}_{\pmb{\theta}}$ is an upper bound of $\mathcal{I}_{\pmb{\theta}}$. In both cases, we can use $\mathcal{L}_{\pmb{\theta}}$ as a surrogate objective to minimize Eq.~\ref{eq:mutual}. However, note that computing the two expectations in Eq.~4 or Eq.~5 is generally intractable, but one can estimate these two quantities by using Monte Carlo, for instance by sampling first from $Q_1$ and then sampling from $Q_2$. Also, note that the conditions on $Q_1$ and $Q_2$ in the theorem can be satisfied by defining $Q_2(\pmb{x}')=Q_{\pmb{\theta}_0}(\pmb{x}'|\pmb{x})$, where $Q_{\pmb{\theta}_0}(\pmb{x}'|\pmb{x})$ is the proposal distribution with parameter vector $\pmb{\theta}_0$ and $Q_1(\pmb{x})=\pi U_\mathcal{X}(\pmb{x}) + (1-\pi) \delta(\pmb{x}-\hat{\pmb{x}})$, where $U$ is a uniform distribution over the whole space $\mathcal{X}$, $\delta$ is the Dirac delta function, $\hat{\pmb{x}}$ is the last accepted sample and $\pi$ defines the proportions of the mixture. Once $\mathcal{L}_{\pmb{\theta}}$ is estimated, we can update the parameters of the two parametrizations by using an off-the-shelf gradient-based optimizer.

In our training algorithm, we choose case (b) as our surrogate objective to minimize the amount of target likelihood evaluations. Indeed, note that the estimate for case (a) requires $O(d^2)$ evaluations, because we need to compute $Q(\pmb{x}''|\pmb{x})$ for all $\pmb{x}''\in N(\pmb{x})$, whereas the one for case (b) requires only $O(d)$ evaluations. Finally, we learn the additional parameter $\eta$ in case (b) using standard gradient descent. The whole learning procedure is shown in Algorithm 1.

\begin{algorithm}[t]
   \caption{Fast Local Self-Balancing (FLSB).}
   \label{alg:flsb}
\begin{algorithmic}
   \STATE {\bfseries Input:} Learning rate $\gamma=1e-2$, $\pi=1e-8$, initial parameter $\pmb{\theta}_0$, burn-in iterations $K$ and batch of samples $N$
   \STATE $\{\hat{\pmb{x}}^{(i)}\}_{i=1}^N\sim U_{\mathcal{X}}$
   \FOR{$k=1$ {\bfseries to} $K$}
    \STATE $\{\pmb{x}^{(i)}\}_{i=1}^N\sim Q_1$
    \STATE $\{\pmb{x}'^{(i)}\}_{i=1}^N\sim Q_2$ (using the approximation)
    \STATE $\widehat{\mathcal{L}}_{\pmb{\theta}}\leftarrow$ Estimate $\mathcal{L}_{\pmb{\theta}}$ using $\{\pmb{x}^{(i)}\}_{i=1}^N$, $\{\pmb{x}'^{(i)}\}_{i=1}^N$
    \STATE $\pmb{\theta} \leftarrow \pmb{\theta}-\frac{\gamma}{N}\nabla_{\pmb{\theta}}\widehat{\mathcal{L}}_{\pmb{\theta}}$
    \STATE Update $\eta$
    \STATE Accept/Reject samples (exact accept. score)
    \STATE Update $\{\hat{\pmb{x}}^{(i)}\}_{i=1}^N$ with accepted samples
    \STATE $\pmb{\theta}_0\leftarrow\pmb{\theta}$
   \ENDFOR
\end{algorithmic}
\end{algorithm}

\subsection{Constant Target Evaluation}
Without loss of generality, we can express $\tilde{p}(\pmb{x})=e^{f(\pmb{x})}$ for some $f:\mathbb{R}^d\rightarrow\mathbb{R}$ and consider $\mathcal{X}=\{0,1\}^d$.\footnote{$\tilde{p}(\pmb{x})$ is obviously defined over a discrete sample space. However, we can always identify a real-valued function which coincides with $\log\tilde{p}(\pmb{x})$ on the discrete support.} By definition of $N(\pmb{x})$, the proposal distribution in Eq.~\ref{eq:proposal} can be written in an equivalent form as $Q(\pmb{x}'|\pmb{x})=\sum_{i=1}^d Q(\pmb{x}'|\pmb{x},i)Q(i|\pmb{x})$, where $Q(\pmb{x}'|\pmb{x},i)=\delta(\pmb{x}'-\pmb{x}_{-i})$ with $\pmb{x}_{-i}$ obtained by flipping $i-$bit in $\pmb{x}$, and
\begin{align}
    Q(i|\pmb{x})=\text{Cat}\Big\{\text{Norm}\Big[g\big(e^{df_1(\pmb{x})},\dots,g\big(e^{df_d(\pmb{x})}\big)\Big]\Big\}_i \nonumber
\end{align}
where $\text{Cat}$ stands for a categorical distribution, $\text{Norm}$ is a normalization operator acting on a $d-$dimensional vector and $df_i(\pmb{x})=f(\pmb{x}_{-i})-f(\pmb{x})$. It's clear that computing previous equation requires to evaluate $f$ for $O(d)$ times. However, if we assume that $f$ is known and differentiable (which is true for instance in energy-based models), we can use Taylor expansion to approximate the difference $df_i(\pmb{x})$. Indeed, we have that
\begin{align}
    f(\pmb{x}_{-i})-f(\pmb{x})\approx \nabla_{\bar{\pmb{x}}}f(\bar{\pmb{x}})|^T_{\bar{\pmb{x}}=\pmb{x}}(\pmb{x}_{-i}-\pmb{x})\doteq\bar{df}_i(\pmb{x}) \nonumber
\end{align}
which allows to evaluate the target only $O(1)$ times. Therefore, our new proposal distribution is defined as follows:
\begin{align}
    Q(i|\pmb{x})=\text{Cat}\Big\{\text{Norm}\Big[g\big(e^{\bar{df}_1(\pmb{x})},\dots,g\big(e^{\bar{df}_d(\pmb{x})}\big)\Big]\Big\}_i \nonumber
\end{align}
Interestingly, if we choose $g(t)=\sqrt{t}$, we recover the exact same proposal distribution of a recent sampler, called Gibbs-With-Gradients~\cite{grathwohl2021oops}. Thanks to this approximation, we can propose a more efficient version of Algorithm 1, called Fast Local Self-Balancing procedure (FLSB), shown in Algorithm 2.

Finally, we can summarize the main differences among locally balanced proposals of~\cite{zanella2020informed}, Gibbs-With-Gradients~\cite{grathwohl2021oops} and our samplers in Table~\ref{tab:summary}.

\begin{table}[t]
\caption{Summary comparing locally balanced proposals (LB) of~\cite{zanella2020informed}, Gibbs-With-Gradients (GWG)~\cite{grathwohl2021oops} and our samplers (LSB and FLSB).} \label{sample-table}
\label{tab:summary}
\begin{center}
\begin{tabularx}{0.6\textwidth}{llll}
\textbf{Name}  & $g(t)$ & $df_i$ & \textbf{Eval./step}  \\
\hline
LB & Fixed & Exact & $O(d)$ \\
GWG & Fixed ($g(t)=\sqrt{t}$) & Approx.* & $O(1)$ \\
LSB & Learnt & Exact & $O(d)$ \\
FLSB & Learnt & Approx.* & $O(1)$ \\
\hline
\multicolumn{4}{l}{* Assumption: $f$ is known and differentiable.}
\end{tabularx}
\end{center}
\end{table}

{\section{Related Work}
	\label{sec:related}}
It's important to devise strategies, which enable the automatic adaption  of proposals to target distributions, not only to reduce user intervention, but also to increase the efficiency of MCMC samplers~\cite{andrieu2008tutorial,hoffman2014no}. Recently, there has been a surge of interest in using machine learning and in particular deep learning to learn proposals directly from data, especially in the continuous domain. Here, we provide a brief overview of recent integrations of machine learning and MCMC samplers according to different parametrizations and training objectives.

\textbf{Parametrizations and objectives in the continuous domain}. The work in~\cite{wang2018meta} proposes a strategy based on block Gibbs sampling, where blocks are large motifs of the underlying probabilistic graphical structure. It parameterizes the conditional distributions of each block using mixture density networks and trains them using meta-learning on a log-likelihood-based objective. The work in~\cite{song2017nice} considers a global sampling strategy, where the proposal is parameterized by a deep generative model. The model is learnt through adversarial training, where a neural discriminator is used to detect whether or not generated samples are distributed according to the target distribution. Authors in~\cite{habib2018auxiliary} propose a global sampling strategy based on MCMC with auxiliary variables~\cite{higdon1998auxiliary}. The proposals are modelled as Gaussian distributions parameterized by neural networks and are trained on a variational bound of a log-likelihood-based objective. The works in~\cite{levy2018generalizing,gong2019meta} propose a gradient-based MCMC~\cite{duane1987hybrid,grenander1994representations}, where neural models are used to learn the hyperparameters of the equations governing the dynamics of the sampler. Different objectives are used during training. In particular, the work in~\cite{gong2019meta} uses a log-likelihood based objective, whereas the work in~\cite{levy2018generalizing} considers the expected squared jump distance, namely a tractable proxy for the lag-1 autocorrelation function~\cite{pasarica2010adaptively}. The work in~\cite{zhu2019sample} proposes a global two-stage strategy, which consists of (i) sampling according to a Gaussian proposal and (ii) updating its parameters using the first- and second-order statistics computed from a properly maintained pool of samples. The parameter update can be equivalently seen as finding the solution maximizing a log-likelihood function defined over the pool of samples. Finally, the work in~\cite{pompe2020framework} extends this last strategy to the case of Gaussian mixture proposals. All these works differ from the current one in at least two aspects. Firstly, it is not clear how these parametrizations  can be applied to sampling in the discrete domain. Secondly, the proposed objectives compute either a distance between the proposal distribution and the target one, namely using an adversarial objective or a variational bound on the log-likelihood, or a proxy on the correlation between consecutive generated samples, namely the expected squared jump distance. Instead, our proposed objective is more general in the sense that it reduces the statistical dependence between consecutive samples, as being closely related to mutual information.

\textbf{Sampling in the discrete domain}. Less efforts have been devoted to devise sampling strategies for a purely discrete domain. Most of the works consider problem relaxations by embedding the discrete domain into a continuous one, applying existing strategies like Hamiltonian Monte Carlo~\cite{zhang2012continuous,pakman2013auxiliary,afshar2015reflection,dinh2017probabilistic,nishimura2020discontinuous} on it and then moving back to the original domain. These strategies are suboptimal, either because they consider limited settings, where the target distribution has specific analytic forms~\cite{zhang2012continuous}, or because they make strong assumptions on the properties of the embeddings, thus not preserving the topological properties of the discrete domain~\cite{pakman2013auxiliary,afshar2015reflection,dinh2017probabilistic,nishimura2020discontinuous}.\footnote{For example by considering transformations that are bijective and/or by proposing transformations which allow to tractably compute the marginal distribution on the continuous domain.}. The work in~\cite{zanella2020informed} provides an extensive experimental comparison between several discrete sampling strategies, including the ones based on embeddings, based on stochastic local search~\cite{hans2007shotgun} and the Hamming ball sampler~\cite{titsias2017hamming}, which can be regarded as a more efficient version of block Gibbs sampling. Notably, the sampling strategy based on locally informed proposals and balancing functions proposed in~\cite{zanella2020informed} can be considered as the current state of the art for discrete MCMC. Our work builds and extends upon this sampler by integrating it with a machine learning strategy. 

It's important to mention that there are a couple of neural approaches applied to the discrete domain. The first one~\cite{jaini2021sampling} proposes to use normalizing flows to (i) learn a continuous relaxation of the discrete domain by dequantizing input data, and to (ii) learn a latent embedding more amenable to MCMC sampling. Learning the input-latent transformation is performed by maximizing the log-likelihood computed on data sampled from the latent space. The second one~\cite{dai2020learning} proposes a strategy to learn an initializing distribution for a fixed local discrete MCMC sampler in the context of energy-based models. This is achieved by forcing the distribution to be close enough (in terms of KL divergence) to the fixed point of the MCMC kernel. Both works differ from ours in at least three aspects. Indeed, our work parameterizes the kernel of a local MCMC sampler, while the others consider a more global approach. Secondly, we are learning a one-dimensional real valued function through a simple neural network, instead of learning a more complex deep latent variable model which must transform input data. Finally, we are providing a mutual information objective, which directly tackle the problem of reducing the statistical dependence, and therefore also correlation, of consecutive samples.

{\section{Experiments}
	\label{sec:experiments}}
Firstly, we analyze samplers' performance on energy-based models, including the 2D Ising model and Restricted Boltzmann Machines. Then, we perform experiments on additional UAI benchmarks. Code to replicate the experiments is available at \url{https://github.com/emsansone/LSB.git}.

\begin{figure}
     \centering
     \begin{subfigure}[b]{0.45\linewidth}
         \centering
         \includegraphics[width=0.4\linewidth]{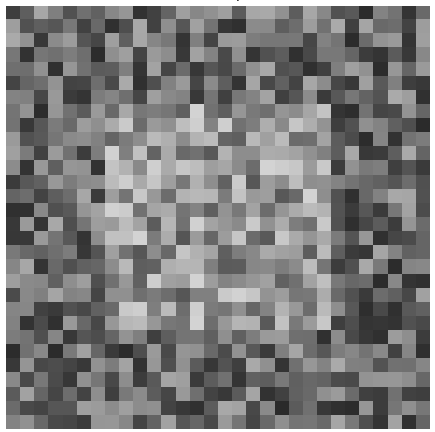}
         \caption{Noisy}
         \label{fig:noisy}
     \end{subfigure}%
     \begin{subfigure}[b]{0.45\linewidth}
         \centering
         \includegraphics[width=0.4\linewidth]{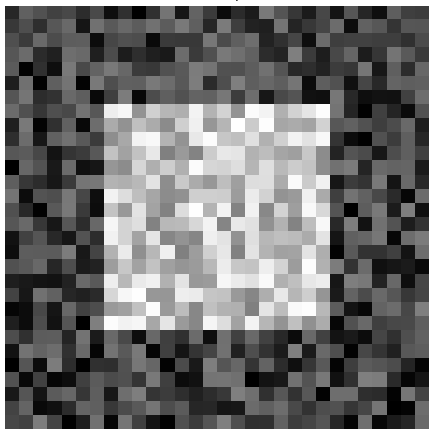}
         \caption{Clean}
         \label{fig:clean}
     \end{subfigure}
     \caption{Examples of $\pmb{\alpha}$ in different settings of the Ising model (30$\times$ 30), i.e noisy $\mu=1,\sigma=3$ and clean $\mu=3,\sigma=3$.}
     \label{fig:alpha}
\end{figure}

\subsection{2D Ising Model}
We consider the Ising model applied to image segmentation, to identify an object from its background. Consider a binary state space $\mathcal{X}=\{-1,1\}^V$, where $(V,E)$ defines a square lattice graph of the same size of the analyzed image, namely $n\times n$. For each state configuration $\mathbf{x}=(x_i)_{i\in V}\in\mathcal{X}$, define a prior distribution 
\begin{align*}
p_{prior}(\mathbf{x})\propto\exp\bigg\{\lambda\sum_{(i,j)\in E}x_i x_j\bigg\}
\end{align*}
where $\lambda$ is a non-negative scalar used to weight the dependence among neighboring variables in the lattice.  Then, consider that each pixel $y_i$ is influenced only by the corresponding hidden variable $x_i$ and generated according to a Gaussian density with mean $\mu x_i$ and variance $\sigma^2$. Note that each variable in the lattice tells whether the corresponding pixel belongs to the object or to the background (1 or -1, respectively). The corresponding posterior distribution of a hidden state $\mathbf{x}$ given an observed image is defined as follows:
\begin{equation}\label{eq:ising}
  p(\mathbf{x})=\frac{1}{Z}\exp\bigg\{\sum_{i\in V}\alpha_i x_i+\lambda\sum_{(i,j)\in E}x_i x_j\bigg\}
\end{equation}
where $\alpha_i=y_i\mu/\sigma^2$ is a coefficient biasing $x_i$ towards either $1$ or $-1$. Therefore, $\pmb{\alpha}=(\alpha_i)_{i\in V}$ contains information about the observed image. Figure~\ref{fig:alpha} shows two synthetically generated examples of $\pmb{\alpha}$. We evaluate the sampling performance on the distribution defined in Eq.~\ref{eq:ising}. Importantly, the topological graph structure of the lattice and the exponential form of the posterior distribution allows to compute a locally balanced proposals in $O(1)$ target evaluations, without the need of using a gradient-based approximation (cf. Section 3.3). Therefore, we consider only comparisons between LB and LSB
\begin{figure*}
     \centering
     \begin{subfigure}[b]{0.08\textwidth}
         \centering
         \includegraphics[width=0.9\textwidth]{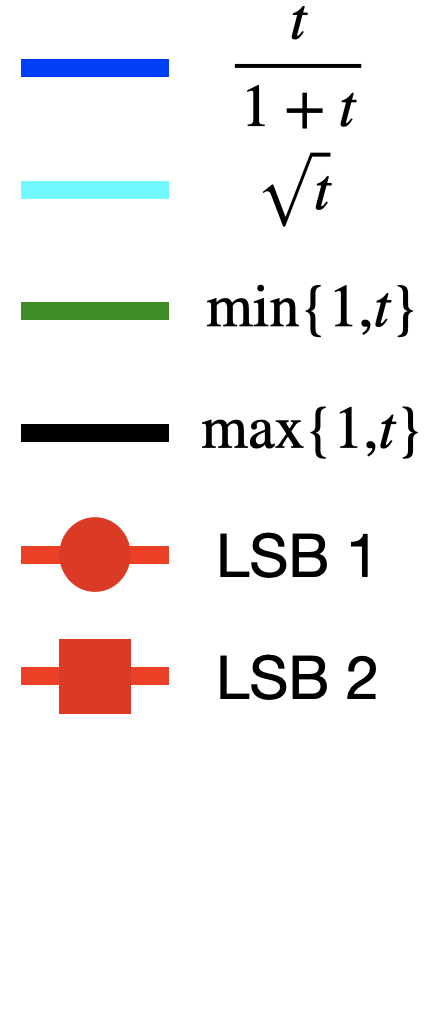}
     \end{subfigure}%
     \begin{subfigure}[b]{0.22\textwidth}
         \centering
         \includegraphics[width=0.9\textwidth]{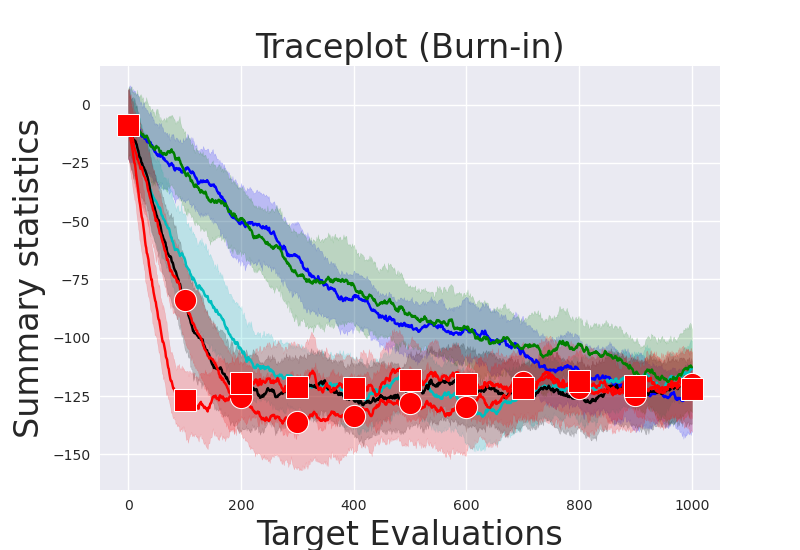}
         \caption{Case 1}
     \end{subfigure}%
     \begin{subfigure}[b]{0.22\textwidth}
         \centering
         \includegraphics[width=0.9\textwidth]{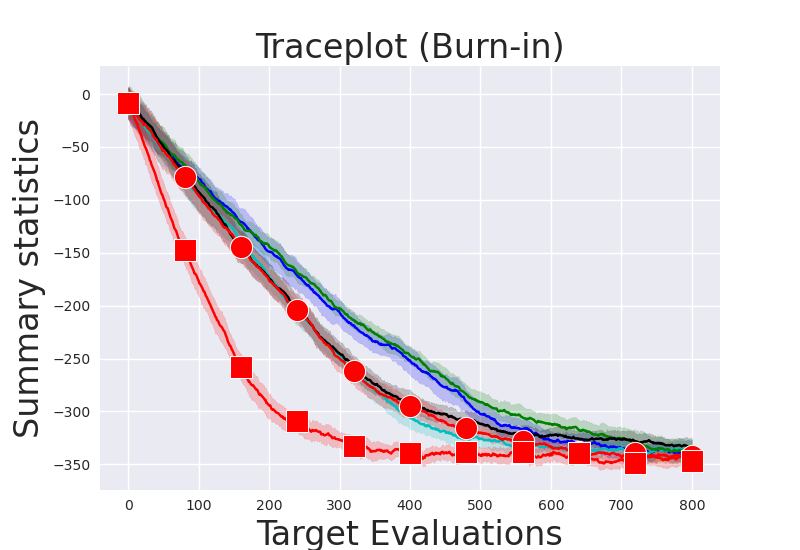}
         \caption{Case 2}
     \end{subfigure}%
     \begin{subfigure}[b]{0.22\textwidth}
         \centering
         \includegraphics[width=0.9\textwidth]{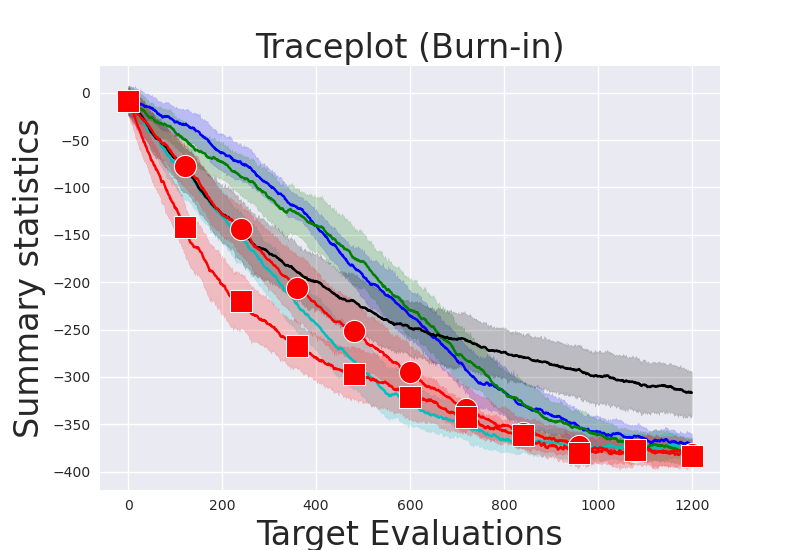}
         \caption{Case 3}
     \end{subfigure}%
     \begin{subfigure}[b]{0.22\textwidth}
         \centering
         \includegraphics[width=0.9\textwidth]{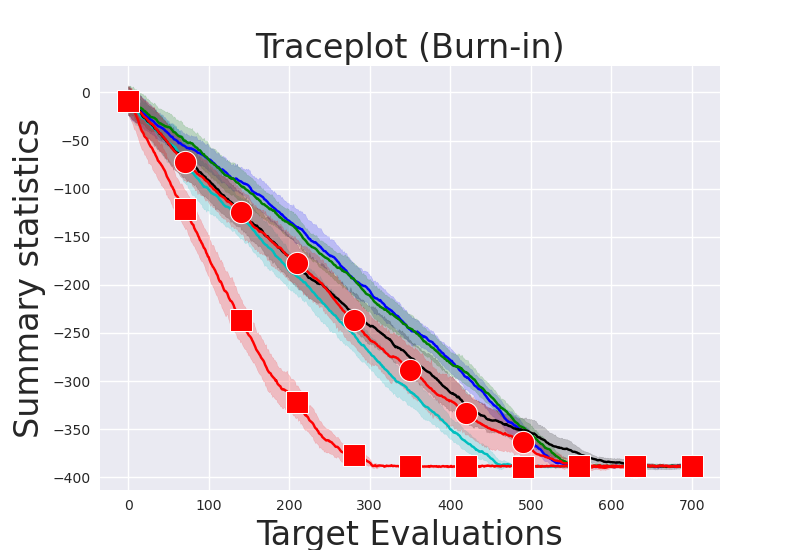}
         \caption{Case 4}
     \end{subfigure}%
     \caption{Samplers' performance on four cases of the Ising model ($30\times 30$) for the burn-in phase. (a) Case 1: Independent-noisy, (b) case 2: Independent-clean, (c) case 3: Dependent-noisy, (d) case 4: Dependent-clean}
     \label{fig:learning}
\end{figure*}
\begin{table*}
 \caption{Quantitative performance for mixing measured by effective sample size on the four cases of the Ising model ($30\times 30$). $\max\{1,t\}$ is performing significantly worse in statistical terms than the other functions.}
 \label{tab:learning}
 \centering
\begin{small}
\begin{tabular}{lcccccc}
    \toprule
    Setting & $\frac{t}{1+t}$ & $\sqrt{t}$ & $\min\{1,t\}$ & $\max\{1,t\}$ & LSB 1 & LSB 2 \\
    \midrule
    Case 1 & $\pmb{2.55\pm 0.30}$ & $\pmb{2.29\pm 0.23}$ & $\pmb{2.43\pm 0.22}$ & $1.71\pm 0.13$ & $\pmb{2.32\pm 0.27}$ & $\pmb{2.34\pm 0.23}$\\
    Case 2 & $\pmb{3.30\pm 0.36}$ & $\pmb{2.89\pm 0.28}$ & $\pmb{2.96\pm 0.30}$ & $1.68\pm 0.11$ & $\pmb{2.85\pm 0.28}$ & $2.30\pm 0.29$ \\
    Case 3 & $\pmb{2.39\pm 0.98}$ & $\pmb{1.83\pm 0.56}$ & $\pmb{2.31\pm 0.92}$ & $1.20\pm 0.10$ & $\pmb{2.01\pm 0.70}$ & $\pmb{2.44\pm 0.96}$\\
    Case 4 & $2.10\pm 0.91$ & $7.08\pm 4.07$ & $1.74\pm 0.26$ & $1.74\pm 0.51$ & $2.52\pm 1.11$ & $\pmb{20.11\pm 15.58}$\\
    \bottomrule
 \end{tabular}
 \end{small}
\end{table*}
\begin{figure}
     \centering
     \begin{subfigure}[b]{0.49\linewidth}
         \centering
         \includegraphics[width=0.9\linewidth]{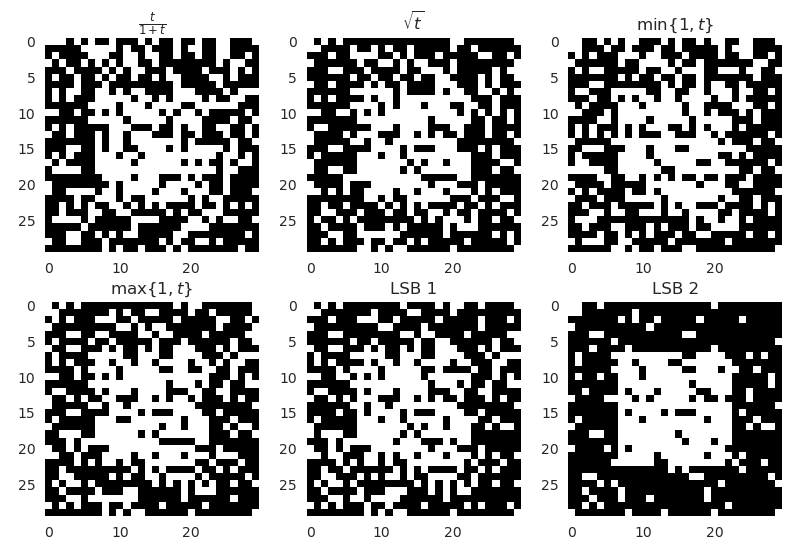}
         \caption{Case 2}
         \label{fig:indep_noisy}
     \end{subfigure}
     \begin{subfigure}[b]{0.49\linewidth}
         \centering
         \includegraphics[width=0.9\linewidth]{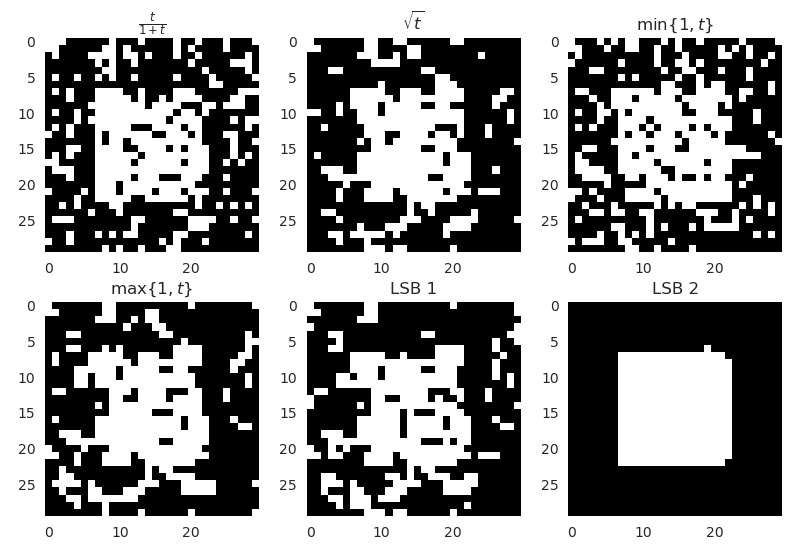}
         \caption{Case 4}
         \label{fig:indep_clean}
     \end{subfigure}%
     \caption{Realizations obtained after $300$ burn-in iterations on the Ising model.}
     \label{fig:img}
\end{figure}

\paragraph{Learning the balancing function.} 
We consider the balancing functions proposed in~\cite{zanella2020informed}, namely $g(t)=t/(1+t)$ (a.k.a Barker function), $\sqrt{t}, \min\{1,t\}$ and $\max\{1,t\}$.\footnote{As discussed in Section 3.3, the case $g(t)=\sqrt{t}$ corresponds to the exact version of GWG. Here there is no need for using the approximation, as we can exploit the structure of the lattice and the exponential form of the distribution to achieve constant target evaluation.}
We compare these four balancing functions with our two parametrizations on four different settings of the Ising model, namely independent and noisy $(\lambda,\mu,\sigma)=(0,1,3)$, independent and clean $(\lambda,\mu,\sigma)=(0,3,3)$, dependent and noisy $(\lambda,\mu,\sigma)=(1,1,3)$ and dependent and clean $(\lambda,\mu,\sigma)=(1,3,3)$ cases and show the corresponding performance in Figure~\ref{fig:learning} and Table~\ref{tab:learning}. We leave additional details and results to the Appendices~\ref{sec:D} and~\ref{sec:E}.

From Figure~\ref{fig:learning}, we can see that our first parametrization LSB 1 is able to always "select" an unbounded balancing function during burn-in, while when approaching convergence it is able to adapt to preserve fast mixing, as measured by the effective sample size (ESS) in Table~\ref{tab:learning}. It's interesting to mention also that the softmax nonlinearity used in LSB 1 can sometimes slow down the adaptation due to vanishing gradients. This can be observed by looking at the case 4 of Figure~\ref{fig:learning}, where for a large part of the burn-in period the strategy prefers $\max\{1,t\}$ over $\sqrt{t}$. Nevertheless, it is still able to recover a solution different from $\max\{1,t\}$ at the end of burn-in, as confirmed by the larger effective sample size in Table~\ref{tab:learning} compared to the one achieved by $\max\{1,t\}$.

Furthermore, we observe that our second parametrization LSB 2, which is functionally more expressive compared to LSB 1, allows to outperform all previous cases in terms of number of target evaluations required to converge, as shown in Figure~\ref{fig:learning} and Table~\ref{tab:learning}. This provides evidence that the optimality of the balancing function depends on the target distribution and that exploiting information about the target can lead to significant improvements (e.g. in case 3 of Figure~\ref{fig:learning}, LSB 2 converges twice time faster as the best balancing function $\sqrt{t}$).
Figure~\ref{fig:img} provides some realizations obtained by the samplers for the cases with dependent variables $\lambda=1$. We clearly see from these pictures that convergence for LSB 2 occurs at an earlier stage than the other balancing functions and therefore the latent variables in the Ising model converge faster to their ground truth configuration.

\subsection{Sampling in Restricted Boltzmann Machines}
We also evaluate the performance on the challenging task of sampling from Restricted Boltzmann Machines, using the experimental setup of~\cite{grathwohl2021oops}. In particular, we train a RBM with $250$ hidden units on the binary MNIST dataset using contrastive divergence and use this model as our base distribution for evaluating the samplers. Notably, since the distribution has an analytic differentiable form, we can exploit the gradient-based approximation explained in Section 3.3. Therefore, we compare our sampler (FLSB) against Gibbs-With-Gradients (GWG)~\cite{grathwohl2021oops}, block Gibbs sampling (Gibbs 2) and the Hamming Ball sampler (HB-10-1)~\cite{titsias2017hamming}.\footnote{Gibbs-X refers to Gibbs sampling with block size of X, whereas HB-X-Y refers to Hamming Ball sampler with block size of X and a Hamming ball of size Y.}  We use two scores to measure the performance. The first one consists of the maximum mean discrepancy (MMD) distance between the generated samples and the ground truth ones, obtained through the block-Gibbs sampling procedure available in RBMs. The second one consists of the effective sample size (ESS) of a summary statistics computed on the sampling trajectories (more details about the experiments are available in Appendix~\ref{sec:D}). We also provide the effective sample size normalized over time (ESS/sec).

\begin{figure}[t]
     \centering
     \begin{subfigure}[b]{0.43\linewidth}
         \centering
         \includegraphics[width=0.9\linewidth]{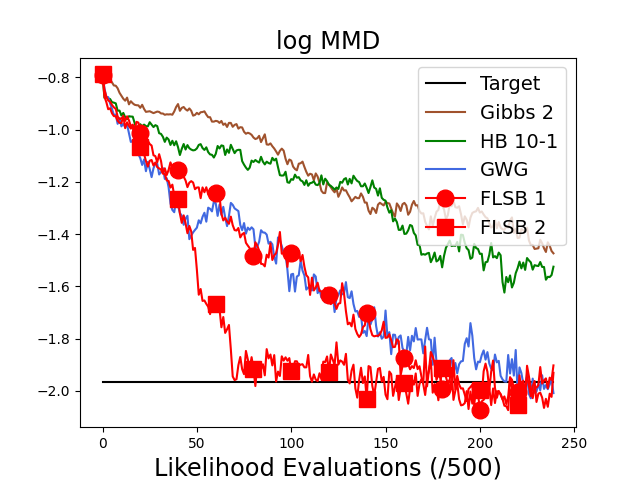}
     \end{subfigure} %
     \begin{subfigure}[b]{0.43\linewidth}
         \centering
         \includegraphics[width=0.9\linewidth]{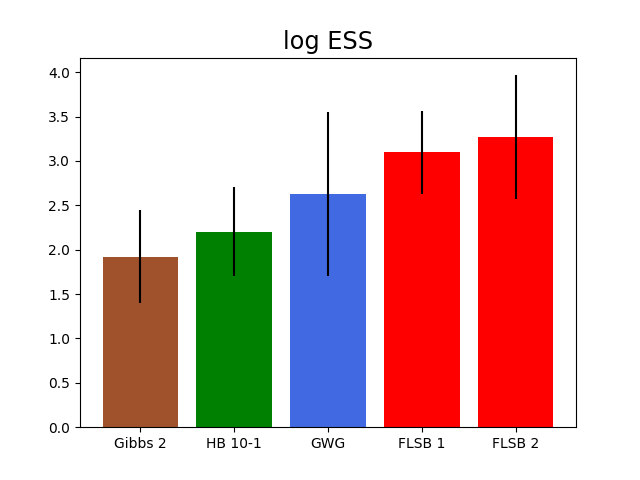}
     \end{subfigure}%
     \begin{subfigure}[b]{0.43\linewidth}
         \centering
         \includegraphics[width=0.9\linewidth]{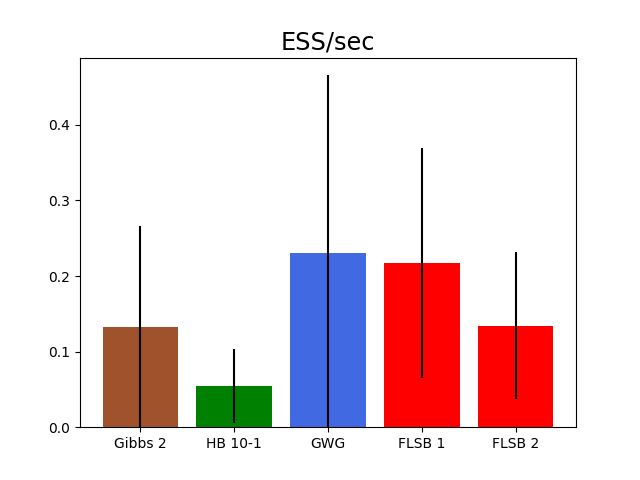}
     \end{subfigure}%
     \caption{Samplers' performance on RBMs. On the top, burn-in performance computed using MMD (in logarithmic scale). On the bottom, mixing performance computed after burn-in using $\log(ESS)$ (on the left) and ESS per seconds (on the right).}
     \label{fig:rbm}
\end{figure}

In Figure~\ref{fig:rbm}, we see that FLSB 1 recovers the performance of GWG. FLSB 2 is able to adapt to the target distribution and to converge using a much smaller number of target likelihood evaluations. Furthermore, we observe that FLSB 2 performs well also in terms of ESS. These experiments provide further evidence that there is a clear advantage on learning the balancing function. However, it is important to mention that the improved performance comes with a computational overhead. Indeed, when comparing the samplers based on ESS/sec (Figure~\ref{fig:rbm}), we observe that, while FLSB 1 and GWG achieve comparable results, the performance of FLSB 2 are inferior on average. This is explained by the fact that each sampling iteration requires to evaluate a more complex balancing function (i.e. a multilayer perceptron network with one 10-neuron hidden layer, corresponding to 31 parameters vs. 4 parameters for FLSB 1). Clearly, there is a distinction between the number of evaluations of the likelihood and the number of evaluations for the balancing function. Our proposed strategy reduces the first ones and the introduced computational overhead affects only the second kind of evaluations. This can be mitigated for example by looking at more efficient block-wise implementations.
However, this is left to future work. Additional comparisons of FLSB2 against block-wise sampling strategies together with a summary of its benefits are provided in Appendix~\ref{sec:G}.
\begin{table*}
 \caption{Quantitative performance for mixing measured by effective sample size on two Markov networks from UAI competition.}
 \label{tab:bayes}
 \centering
\begin{small}
  \begin{tabular}{lcccccc}
    \toprule
    Dataset & $\frac{t}{1+t}$ & $\sqrt{t}$ & $\min\{1,t\}$ & $\max\{1,t\}$ & LSB 1 & LSB 2 \\
    \midrule
    MN 1 & $\pmb{2.90\pm 0.76}$ & $\pmb{3.41\pm 0.77}$ & $2.54\pm 0.32$ & $\pmb{2.70\pm 0.63}$ & $\pmb{3.19\pm 0.46}$ & $\pmb{3.22\pm 0.38}$\\
    MN 2 & $\pmb{3.43\pm 0.75}$ & $\pmb{3.92\pm 0.94}$ & $\pmb{3.78\pm 0.50}$ & $\pmb{3.63\pm 0.67}$ & $\pmb{3.52\pm 0.42}$ & $\pmb{3.44\pm 0.44}$ \\
\bottomrule
 \end{tabular}
\end{small}
\end{table*}
\begin{figure}[t]
     \centering
     \begin{subfigure}[b]{0.08\textwidth}
         \centering
         \includegraphics[width=0.9\textwidth]{img/LEGEND.png}
     \end{subfigure}%
     \begin{subfigure}[b]{0.22\textwidth}
         \centering
         \includegraphics[width=0.9\textwidth]{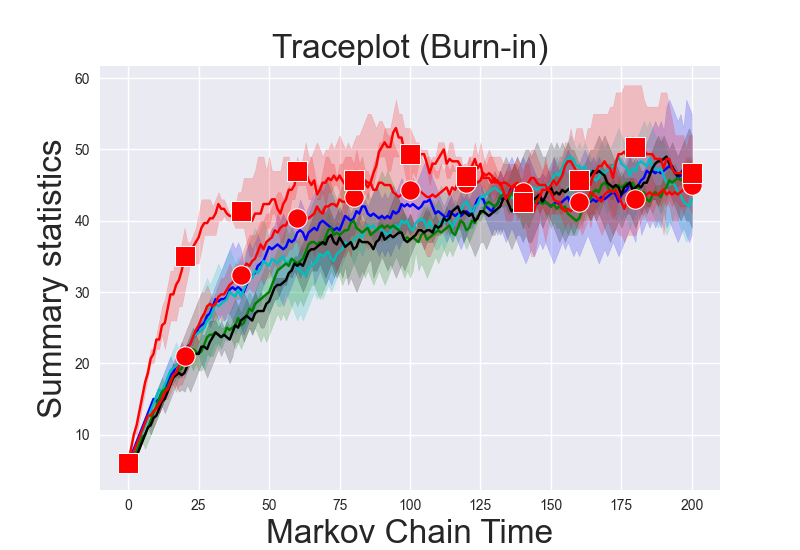}
         \caption{MN 1}
     \end{subfigure}%
     \begin{subfigure}[b]{0.22\textwidth}
         \centering
         \includegraphics[width=0.9\textwidth]{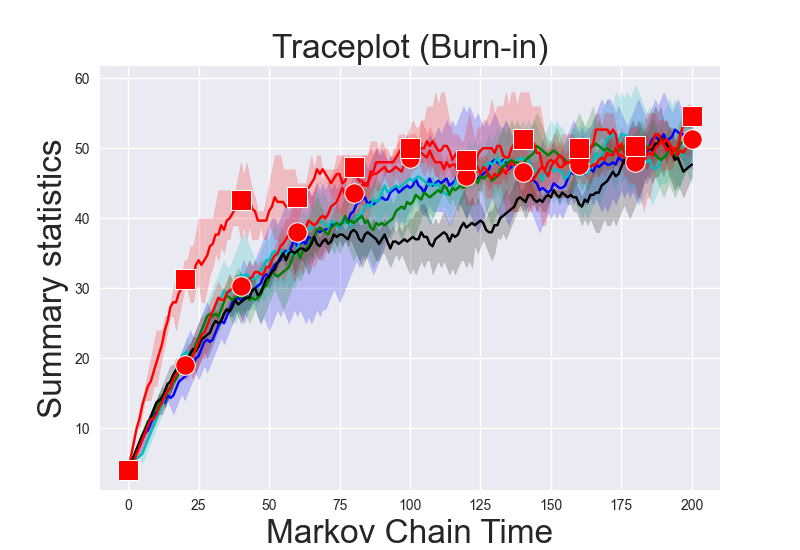}
         \caption{MN 2}
     \end{subfigure}%
    \caption{Samplers' performance on Markov networks from UAI competition ($100$ variables). (a)-(b) are the traceplots for the burn-in phase.}
     \label{fig:bayes}
\end{figure}
\subsection{Markov Networks: UAI data}
Lastly, we evaluate how our strategy generalizes to different graph topologies compared to the one of the Ising model. In particular, we consider two Markov networks, with $100$ discrete variables each and near-deterministic dependencies, from the 2006 UAI competition.\footnote{\url{http://sli.ics.uci.edu/~ihler/uai-data/}} In this setting, we can't leverage the gradient-based approximation of Section 3.3, as the distribution is specified in tabular form. Therefore, we compare our sampler LSB with the four balancing functions proposed in~\cite{zanella2020informed}.  Similarly to previous experiments, we analyze the convergence of the burn-in phase (using traceplots) and the mixing performance according to ESS. Further details about the simulations are available in the Appendices~\ref{sec:D} and~\ref{sec:F}.

Also in this case, we observe that the proposed strategy is able to adapt to the target distribution and reduce the number of target likelihood evaluations required to converge (Figure~\ref{fig:bayes}). In this case, we observe similar performance in terms of ESS (cf.~Table~\ref{tab:bayes}).

{\section{Conclusions and Future Work}
	\label{sec:conclusion}}
We have presented a strategy to learn locally balanced proposals for MCMC in discrete spaces. The strategy consists of (i) a new parametrization of balancing functions and (ii) a learning procedure adapting the proposal to the target distribution. This allows to reduce the number of target likelihood evaluations required to converge. We believe that the proposed strategy can play an important role on applications where querying an oracle distribution is very expensive (like in deep energy-based models). We will investigate this in the future.

Note that the LSB sampler belongs to the family of local sampling strategies, thus inheriting their limitations. The locality assumption can be quite restrictive, for example when sampling from discrete distributions with deterministic dependencies among variables. In such situations, local sampling might fail to correctly sample from the target in a finite amount of time, as being required to cross regions with zero probability mass. The locality assumption can be relaxed for instance by leveraging a recent extension of the works in~\cite{zanella2020informed,grathwohl2021oops}, called Path Auxiliary MCMC~\cite{sun2021path}, which uses a trajectory of samples generated by the repeated application of the same base locally balanced proposal, and by adapting our proposed strategy to this new setting. Alternatively, we can consider block-wise implementations of locally balanced proposals, as also suggested in~\cite{zanella2020informed}. However, we leave all these extensions to future work.

To the best of our knowledge, we are the first to consider using mutual information to accelerate MCMC. We have focused on a local MCMC sampler for discrete domains. In the future, it will be interesting to see the application of the mutual information criterion to more global settings as well as its extension to the continuous case.

Lastly, further theoretical investigation is required to establish the local convergence of our training algorithm. Indeed, standard results of gradient-based optimization do not apply in this setting, as the iterates are correlated. This remains an open challenge for future work.


\section*{Acknowledgements}
This research was partially supported by TAILOR, a project funded by EU Horizon 2020 research and innovation programme under GA No 952215. The author would like to thank Luc de Raedt for his support.

\bibliographystyle{alpha}
\bibliography{main}

\appendix
{\section{Detailed Balance}
	\label{sec:A}}
	
We want to prove that $p(\mathbf{x})T(\mathbf{x}'|\mathbf{x})=p(\mathbf{x}')T(\mathbf{x}|\mathbf{x}')$ for all $\mathbf{x}'\neq\mathbf{x}$.
We have that
\begin{align}
    p(\mathbf{x})&A(\mathbf{x}',\mathbf{x})\frac{g\big(\frac{\tilde{p}(\mathbf{x}')}{\tilde{p}(\mathbf{x})}\big)1[\mathbf{x}'\in N(\mathbf{x})]}{Z(\mathbf{x})} = \nonumber\\ & p(\mathbf{x}')A(\mathbf{x},\mathbf{x}')\frac{g\big(\frac{\tilde{p}(\mathbf{x})}{\tilde{p}(\mathbf{x}')}\big)1[\mathbf{x}\in N(\mathbf{x}')]}{Z(\mathbf{x}')} \nonumber
\end{align}
By observing that $1(\mathbf{x}'\in N(\mathbf{x}))=1(\mathbf{x}\in N(\mathbf{x}'))$ and using the balancing property, we can simplify previous equality to obtain the following relation:
\begin{align}
    p(\mathbf{x})&A(\mathbf{x}',\mathbf{x})\frac{g\big(\frac{\tilde{p}(\mathbf{x}')}{\tilde{p}(\mathbf{x})}\big)}{Z(\mathbf{x})} = \nonumber\\ & p(\mathbf{x}')A(\mathbf{x},\mathbf{x}')\frac{\frac{\tilde{p}(\mathbf{x})}{\tilde{p}(\mathbf{x}')}g\big(\frac{\tilde{p}(\mathbf{x}')}{\tilde{p}(\mathbf{x})}\big)}{Z(\mathbf{x}')} \nonumber
\end{align}
Therefore, we can apply standard algebra to simplify even more
\begin{align}
    \tilde{p}(\mathbf{x})A(\mathbf{x}',\mathbf{x}) = \tilde{p}(\mathbf{x})A(\mathbf{x},\mathbf{x}')\frac{Z(\mathbf{x})}{Z(\mathbf{x}')} \nonumber
\end{align}
Finally, recall that for balancing functions $A(\mathbf{x},\mathbf{x}')=\min\big\{1,\frac{\tilde{p}(\mathbf{x})Q(\mathbf{x}'|\mathbf{x})}{\tilde{p}(\mathbf{x}')Q(\mathbf{x}|\mathbf{x}')}\big\}=\min\big\{1,\frac{Z(\mathbf{x}')}{Z(\mathbf{x})}\big\}$ and therefore previous equality becomes an identity, namely:
\begin{align}
    \tilde{p}(\mathbf{x})A(\mathbf{x}',\mathbf{x}) = \tilde{p}(\mathbf{x})A(\mathbf{x}',\mathbf{x}) \nonumber
\end{align}
thus proving detailed balance.

{\section{Ergodicity}
	\label{sec:B}}
Let's consider a Markov chain, namely
\begin{align}
    T(\mathbf{x}'|\mathbf{x}) =A(\mathbf{x}',\mathbf{x})Q(\mathbf{x}'|\mathbf{x}) + 1[\mathbf{x}'=\mathbf{x}]\sum_{\mathbf{x}''\in\mathcal{X}}\big(1-A(\mathbf{x}'',\mathbf{x})\big)Q(\mathbf{x}''|\mathbf{x}))
    \label{eq:trans}
\end{align}
with a proposal of the following form:
\begin{equation}
    Q(\mathbf{x}'|\mathbf{x}) = \frac{g\big(\frac{\tilde{p}(\mathbf{x}')}{\tilde{p}(\mathbf{x})}\big)1[\mathbf{x}'\in N(\mathbf{x})]}{Z(\mathbf{x})}
    \label{eq:propos}
\end{equation}
We can prove the ergodicity of the Markov chain for the case where the fixed-point distribution $p(\mathbf{x})>0$ for every $\mathbf{x}\in\mathcal{X}$ and then extend it to a general distribution $p$.

Now, assume that $p(\mathbf{x})>0$ for any point $\mathbf{x}\in\mathcal{X}$, $g(t)>0$ for any $t>0$ and $\mathcal{X}$ is a $d$-dimensional discrete space. Then, the Markov chain in Eq.~\ref{eq:trans} with proposal defined according to Eq.~\ref{eq:propos} can reach any state $\mathbf{x}'$ from any state $\mathbf{x}$ in $d$ steps with non-zero probability. More formally, we can construct a new Markov chain by applying $d$ times the original one and identify its transition probability with $T^d(\mathbf{x}'|\mathbf{x})$. We can easily check, thanks to our assumptions, that $T^d(\mathbf{x}'|\mathbf{x})>0$ for any $\mathbf{x},\mathbf{x}'$. In other words, the original Markov chain is regular. This is sufficient to satisfy the assumptions of the fundamental theorem of homogeneous Markov chains~\cite{neal1993probabilistic}, thus proving ergodicity.

We can extend the previous result to any arbitrary $p$ (namely considering cases where $p(\mathbf{x})=0$ for some $\mathbf{x}\in\mathcal{X}$). This can be achieved by modifying our assumptions on $g$, namely considering that $g(t)>0$ for any $t\geq 0$ and reusing the same proof strategy.

%
%
%
%

{\section{Proof of Theorem 1}
	\label{sec:C}}
We want to prove the following theorem.

\begin{theorem}
Consider $\mathcal{X}=\{0,1\}^d$ and $p(\pmb{x})=\tilde{p}(\pmb{x})/\Gamma$, where $\Gamma$ is a normalizing constant. Define
two auxiliary distributions $Q_1$ and $Q_2$, such that for all $\pmb{x}\in\mathcal{X}$, $Q_1(\pmb{x})>0$ whenever $\tilde{p}(\pmb{x})>0$, and for all $\pmb{x}'\in\mathcal{N(\pmb{x})}$, $Q_2(\pmb{x}')>0$ whenever $Q(\pmb{x}'|\pmb{x})>0$.
Therefore,
\begin{itemize}
    \item[(a)] if $\mathcal{M}(\pmb{x})=1-\sum_{\pmb{x}''\in N(\pmb{x})}A(\pmb{x}'',\pmb{x})Q(\pmb{x}''|\pmb{x})$, then
    \begin{align}
        \mathcal{L_{\pmb{\theta}}} =& E_{\substack{\pmb{x}\sim Q_1 \\ \pmb{x}'\sim Q_2}}\bigg\{\frac{\tilde{p}(\pmb{x})Q(\pmb{x}'|\pmb{x})}{Q_1(\pmb{x})Q_2(\pmb{x}')}A(\pmb{x}',\pmb{x})\log\frac{A(\pmb{x}',\pmb{x})Q(\pmb{x}'|\pmb{x})}{\tilde{p}(\pmb{x}')}\bigg\} + E_{\pmb{x}\sim Q_1}\bigg\{\frac{\tilde{p}(\pmb{x})}{Q_1(\pmb{x})}\mathcal{M}(\pmb{x})\log\frac{\mathcal{M}(\pmb{x})}{\tilde{p}(\pmb{x})}\bigg\}
    \end{align}
    and $\mathcal{I}_{\pmb{\theta}}=\frac{\mathcal{L_{\pmb{\theta}}}}{\Gamma} + \frac{\log\Gamma}{\Gamma}$.
    \item[(b)] if $\mathcal{M}(\pmb{x})=1-A(\pmb{x}^*,\pmb{x})Q(\pmb{x}^*|\pmb{x})$, where $\pmb{x}^*$ is randomly sampled according to a uniform distribution over $N(\pmb{x})$, then
    \begin{align}
        \mathcal{L_{\pmb{\theta}}} =& E_{\substack{\pmb{x}\sim Q_1 \\ \pmb{x}'\sim Q_2}}\bigg\{\frac{\tilde{p}(\pmb{x})Q(\pmb{x}'|\pmb{x})}{Q_1(\pmb{x})Q_2(\pmb{x}')}A(\pmb{x}',\pmb{x})\log\frac{A(\pmb{x}',\pmb{x})Q(\pmb{x}'|\pmb{x})}{\tilde{p}(\pmb{x}')}\bigg\} + E_{\pmb{x}\sim Q_1}\bigg\{\frac{\mathcal{M}(\pmb{x})}{Q_1(\pmb{x})}\bigg[\eta \mathcal{M}(\pmb{x}) - \tilde{p}(\pmb{x})\log\eta  -\tilde{p}(\pmb{x})\bigg]\bigg\}
    \end{align}
    and $\mathcal{I}_{\pmb{\theta}}\leq\frac{\mathcal{L_{\pmb{\theta}}}}{\Gamma} + \frac{\log\Gamma}{\Gamma}$, for $\eta\in\mathbb{R}^+$.
\end{itemize}
\end{theorem}
\begin{proof}
Let's start from the definition of $\mathcal{I}_{\pmb{\theta}}$ and use the fact that $T_{\pmb{\theta}}(\pmb{x}'|\pmb{x})=A(\mathbf{x}',\mathbf{x})Q(\mathbf{x}'|\mathbf{x}) + 1[\mathbf{x}'=\mathbf{x}]\sum_{\mathbf{x}''\in\mathcal{X}}\big(1-A(\mathbf{x}'',\mathbf{x})\big)Q(\mathbf{x}''|\mathbf{x}))$.
\begin{align}
    \mathcal{I}_{\pmb{\theta}} &= KL\big\{p(\pmb{x})T_{\pmb{\theta}}(\pmb{x}'|\pmb{x})\|p(\pmb{x})p(\pmb{x}')\big\} \nonumber\\
    &=\sum_{\pmb{x}\in\mathcal{X}}\sum_{\pmb{x}'\in\mathcal{X}}p(\pmb{x})T_{\pmb{\theta}}(\pmb{x}'|\pmb{x})\log\frac{p(\pmb{x})T_{\pmb{\theta}}(\pmb{x}'|\pmb{x})}{p(\pmb{x})p(\pmb{x}')} \nonumber\\
    &=\sum_{\pmb{x}\in\mathcal{X}}\sum_{\pmb{x}'\in\mathcal{X}}p(\pmb{x})T_{\pmb{\theta}}(\pmb{x}'|\pmb{x})\log\frac{T_{\pmb{\theta}}(\pmb{x}'|\pmb{x})}{p(\pmb{x}')} \nonumber\\
    &=\sum_{\pmb{x}\in\mathcal{X}}\sum_{\pmb{x}'\in N(\pmb{x})\cup \{\pmb{x}\}}p(\pmb{x})T_{\pmb{\theta}}(\pmb{x}'|\pmb{x})\log\frac{T_{\pmb{\theta}}(\pmb{x}'|\pmb{x})}{p(\pmb{x}')} \nonumber\\
    &=\frac{1}{\Gamma}\sum_{\pmb{x}\in\mathcal{X}}\sum_{\pmb{x}'\in N(\pmb{x})\cup \{\pmb{x}\}}\tilde{p}(\pmb{x})T_{\pmb{\theta}}(\pmb{x}'|\pmb{x})\log\frac{T_{\pmb{\theta}}(\pmb{x}'|\pmb{x})}{\tilde{p}(\pmb{x}')} + \frac{\log\Gamma}{\Gamma}\nonumber\\
    &=\frac{1}{\Gamma}\sum_{\pmb{x}\in\mathcal{X}}\sum_{\pmb{x}'\in N(\pmb{x})}\tilde{p}(\pmb{x})T_{\pmb{\theta}}(\pmb{x}'|\pmb{x})\log\frac{T_{\pmb{\theta}}(\pmb{x}'|\pmb{x})}{\tilde{p}(\pmb{x}')} + \frac{1}{\Gamma}\sum_{\pmb{x}\in\mathcal{X}}\tilde{p}(\pmb{x})T_{\pmb{\theta}}(\pmb{x}|\pmb{x})\log\frac{T_{\pmb{\theta}}(\pmb{x}|\pmb{x})}{\tilde{p}(\pmb{x})} + \frac{\log\Gamma}{\Gamma}\nonumber\\
    &=\frac{1}{\Gamma}\sum_{\pmb{x}\in\mathcal{X}}\sum_{\pmb{x}'\in N(\pmb{x})}\tilde{p}(\pmb{x})A(\mathbf{x}',\mathbf{x})Q(\mathbf{x}'|\mathbf{x})\log\frac{A(\mathbf{x}',\mathbf{x})Q(\mathbf{x}'|\mathbf{x})}{\tilde{p}(\pmb{x}')} + \frac{1}{\Gamma}\sum_{\pmb{x}\in\mathcal{X}}\tilde{p}(\pmb{x})T_{\pmb{\theta}}(\pmb{x}|\pmb{x})\log\frac{T_{\pmb{\theta}}(\pmb{x}|\pmb{x})}{\tilde{p}(\pmb{x})} + \frac{\log\Gamma}{\Gamma}\nonumber\\    
    &=\frac{1}{\Gamma}\sum_{\pmb{x}\in\mathcal{X}}\sum_{\pmb{x}'\in N(\pmb{x})}\tilde{p}(\pmb{x})A(\mathbf{x}',\mathbf{x})Q(\mathbf{x}'|\mathbf{x})\log\frac{A(\mathbf{x}',\mathbf{x})Q(\mathbf{x}'|\mathbf{x})}{\tilde{p}(\pmb{x}')} + \nonumber\\
    &\qquad +
    \frac{1}{\Gamma}\sum_{\pmb{x}\in\mathcal{X}}\tilde{p}(\pmb{x})\sum_{\mathbf{x}''\in\mathcal{X}}\big[1-A(\mathbf{x}'',\mathbf{x})\big]Q(\mathbf{x}''|\mathbf{x})\log\frac{\sum_{\mathbf{x}''\in\mathcal{X}}\big[1-A(\mathbf{x}'',\mathbf{x})\big]Q(\mathbf{x}''|\mathbf{x})}{\tilde{p}(\pmb{x})} + \frac{\log\Gamma}{\Gamma}\nonumber\\
    &=\frac{1}{\Gamma}\sum_{\pmb{x}\in\mathcal{X}}\sum_{\pmb{x}'\in N(\pmb{x})}\tilde{p}(\pmb{x})A(\mathbf{x}',\mathbf{x})Q(\mathbf{x}'|\mathbf{x})\log\frac{A(\mathbf{x}',\mathbf{x})Q(\mathbf{x}'|\mathbf{x})}{\tilde{p}(\pmb{x}')} + \nonumber\\
    &\qquad +
    \frac{1}{\Gamma}\sum_{\pmb{x}\in\mathcal{X}}\tilde{p}(\pmb{x})\big[1 -\sum_{\mathbf{x}''\in\mathcal{X}}A(\mathbf{x}'',\mathbf{x})Q(\mathbf{x}''|\mathbf{x})\big]\log\frac{\big[1 -\sum_{\mathbf{x}''\in\mathcal{X}}A(\mathbf{x}'',\mathbf{x})Q(\mathbf{x}''|\mathbf{x})\big]}{\tilde{p}(\pmb{x})} + \frac{\log\Gamma}{\Gamma}\nonumber\\
    &=\frac{1}{\Gamma}\sum_{\pmb{x}\in\mathcal{X}}\sum_{\pmb{x}'\in N(\pmb{x})}
    Q_1(\pmb{x})Q_2(\pmb{x}')\frac{
    \tilde{p}(\pmb{x})Q(\mathbf{x}'|\mathbf{x})}{Q_1(\pmb{x})Q_2(\pmb{x}')}A(\mathbf{x}',\mathbf{x})\log\frac{A(\mathbf{x}',\mathbf{x})Q(\mathbf{x}'|\mathbf{x})}{\tilde{p}(\pmb{x}')} + \nonumber\\
    &\qquad +
    \frac{1}{\Gamma}\sum_{\pmb{x}\in\mathcal{X}}Q_1(\pmb{x})\frac{\tilde{p}(\pmb{x})}{Q_1(\pmb{x})}\big[1 -\sum_{\mathbf{x}''\in\mathcal{X}}A(\mathbf{x}'',\mathbf{x})Q(\mathbf{x}''|\mathbf{x})\big]\log\frac{\big[1 -\sum_{\mathbf{x}''\in\mathcal{X}}A(\mathbf{x}'',\mathbf{x})Q(\mathbf{x}''|\mathbf{x})\big]}{\tilde{p}(\pmb{x})} + \frac{\log\Gamma}{\Gamma} 
    \end{align}
Now, by defining $\mathcal{M}(\pmb{x})=1 -\sum_{\mathbf{x}''\in\mathcal{X}}A(\mathbf{x}'',\mathbf{x})Q(\mathbf{x}''|\mathbf{x})$, we obtain case (a). Indeed, we have that

\begin{align}
    \mathcal{I}_{\pmb{\theta}}
    &=\frac{1}{\Gamma}\sum_{\pmb{x}\in\mathcal{X}}\sum_{\pmb{x}'\in N(\pmb{x})}
    Q_1(\pmb{x})Q_2(\pmb{x}')\frac{
    \tilde{p}(\pmb{x})Q(\mathbf{x}'|\mathbf{x})}{Q_1(\pmb{x})Q_2(\pmb{x}')}A(\mathbf{x}',\mathbf{x})\log\frac{A(\mathbf{x}',\mathbf{x})Q(\mathbf{x}'|\mathbf{x})}{\tilde{p}(\pmb{x}')} + \nonumber\\
    &\qquad +
    \frac{1}{\Gamma}\sum_{\pmb{x}\in\mathcal{X}}Q_1(\pmb{x})\frac{\tilde{p}(\pmb{x})}{Q_1(\pmb{x})}\big[1 -\sum_{\mathbf{x}''\in\mathcal{X}}A(\mathbf{x}'',\mathbf{x})Q(\mathbf{x}''|\mathbf{x})\big]\log\frac{\big[1 -\sum_{\mathbf{x}''\in\mathcal{X}}A(\mathbf{x}'',\mathbf{x})Q(\mathbf{x}''|\mathbf{x})\big]}{\tilde{p}(\pmb{x})} + \frac{\log\Gamma}{\Gamma}\nonumber\\
    &=\frac{1}{\Gamma}\sum_{\pmb{x}\in\mathcal{X}}\sum_{\pmb{x}'\in N(\pmb{x})}
    Q_1(\pmb{x})Q_2(\pmb{x}')\frac{
    \tilde{p}(\pmb{x})Q(\mathbf{x}'|\mathbf{x})}{Q_1(\pmb{x})Q_2(\pmb{x}')}A(\mathbf{x}',\mathbf{x})\log\frac{A(\mathbf{x}',\mathbf{x})Q(\mathbf{x}'|\mathbf{x})}{\tilde{p}(\pmb{x}')} + \nonumber\\
    &\qquad +
    \frac{1}{\Gamma}\sum_{\pmb{x}\in\mathcal{X}}Q_1(\pmb{x})\frac{\tilde{p}(\pmb{x})}{Q_1(\pmb{x})}\mathcal{M}(\pmb{x})\log\frac{\mathcal{M}(\pmb{x})}{\tilde{p}(\pmb{x})} + \frac{\log\Gamma}{\Gamma}\nonumber\\    
    &=\frac{\mathcal{L}_{\pmb{\theta}}}{\Gamma} + \frac{\log\Gamma}{\Gamma}
\end{align}
For case (b), we can explot the relation $\log y \leq \eta y - \log\eta - 1$ for all $y>0$ we have that

\begin{align}
    \mathcal{I}_{\pmb{\theta}}
    &=\frac{1}{\Gamma}\sum_{\pmb{x}\in\mathcal{X}}\sum_{\pmb{x}'\in N(\pmb{x})}
    Q_1(\pmb{x})Q_2(\pmb{x}')\frac{
    \tilde{p}(\pmb{x})Q(\mathbf{x}'|\mathbf{x})}{Q_1(\pmb{x})Q_2(\pmb{x}')}A(\mathbf{x}',\mathbf{x})\log\frac{A(\mathbf{x}',\mathbf{x})Q(\mathbf{x}'|\mathbf{x})}{\tilde{p}(\pmb{x}')} + \nonumber\\
    &\qquad +
    \frac{1}{\Gamma}\sum_{\pmb{x}\in\mathcal{X}}Q_1(\pmb{x})\frac{\tilde{p}(\pmb{x})}{Q_1(\pmb{x})}\big[1 -\sum_{\mathbf{x}''\in\mathcal{X}}A(\mathbf{x}'',\mathbf{x})Q(\mathbf{x}''|\mathbf{x})\big]\log\frac{\big[1 -\sum_{\mathbf{x}''\in\mathcal{X}}A(\mathbf{x}'',\mathbf{x})Q(\mathbf{x}''|\mathbf{x})\big]}{\tilde{p}(\pmb{x})} + \frac{\log\Gamma}{\Gamma}\nonumber\\
    &\leq\frac{1}{\Gamma}\sum_{\pmb{x}\in\mathcal{X}}\sum_{\pmb{x}'\in N(\pmb{x})}
    Q_1(\pmb{x})Q_2(\pmb{x}')\frac{
    \tilde{p}(\pmb{x})Q(\mathbf{x}'|\mathbf{x})}{Q_1(\pmb{x})Q_2(\pmb{x}')}A(\mathbf{x}',\mathbf{x})\log\frac{A(\mathbf{x}',\mathbf{x})Q(\mathbf{x}'|\mathbf{x})}{\tilde{p}(\pmb{x}')} + \nonumber\\
    &\qquad +
    \frac{1}{\Gamma}\sum_{\pmb{x}\in\mathcal{X}}Q_1(\pmb{x})\frac{\tilde{p}(\pmb{x})}{Q_1(\pmb{x})}\big[1 -\sum_{\mathbf{x}''\in\mathcal{X}}A(\mathbf{x}'',\mathbf{x})Q(\mathbf{x}''|\mathbf{x})\big]\bigg\{\eta\frac{\big[1 -\sum_{\mathbf{x}''\in\mathcal{X}}A(\mathbf{x}'',\mathbf{x})Q(\mathbf{x}''|\mathbf{x})\big]}{\tilde{p}(\pmb{x})} -\log\eta -1\bigg\} + \frac{\log\Gamma}{\Gamma}\nonumber\\
    &=\frac{1}{\Gamma}\sum_{\pmb{x}\in\mathcal{X}}\sum_{\pmb{x}'\in N(\pmb{x})}
    Q_1(\pmb{x})Q_2(\pmb{x}')\frac{
    \tilde{p}(\pmb{x})Q(\mathbf{x}'|\mathbf{x})}{Q_1(\pmb{x})Q_2(\pmb{x}')}A(\mathbf{x}',\mathbf{x})\log\frac{A(\mathbf{x}',\mathbf{x})Q(\mathbf{x}'|\mathbf{x})}{\tilde{p}(\pmb{x}')} + \nonumber\\
    &\qquad +
    \frac{1}{\Gamma}\sum_{\pmb{x}\in\mathcal{X}}Q_1(\pmb{x})\frac{\big[1 -\sum_{\mathbf{x}''\in\mathcal{X}}A(\mathbf{x}'',\mathbf{x})Q(\mathbf{x}''|\mathbf{x})\big]}{Q_1(\pmb{x})}\bigg\{\eta\big[1 -\sum_{\mathbf{x}''\in\mathcal{X}}A(\mathbf{x}'',\mathbf{x})Q(\mathbf{x}''|\mathbf{x})\big] -\tilde{p}(\pmb{x})\log\eta -\tilde{p}(\pmb{x})\bigg\} + \frac{\log\Gamma}{\Gamma}
\end{align}
Finally, we observe that $1-A(\tilde{\pmb{x}},\pmb{x})Q(\tilde{\pmb{x}}|\pmb{x})\geq1 -\sum_{\mathbf{x}''\in\mathcal{X}}A(\mathbf{x}'',\mathbf{x})Q(\mathbf{x}''|\mathbf{x})$ for all $\tilde{\pmb{x}}\in N(\pmb{x})$ and for all $\pmb{x}\in\mathcal{X}$. Therefore,
\begin{align}
    \mathcal{I}_{\pmb{\theta}}
    &\leq\frac{1}{\Gamma}\sum_{\pmb{x}\in\mathcal{X}}\sum_{\pmb{x}'\in N(\pmb{x})}
    Q_1(\pmb{x})Q_2(\pmb{x}')\frac{
    \tilde{p}(\pmb{x})Q(\mathbf{x}'|\mathbf{x})}{Q_1(\pmb{x})Q_2(\pmb{x}')}A(\mathbf{x}',\mathbf{x})\log\frac{A(\mathbf{x}',\mathbf{x})Q(\mathbf{x}'|\mathbf{x})}{\tilde{p}(\pmb{x}')} + \nonumber\\
    &\qquad +
    \frac{1}{\Gamma}\sum_{\pmb{x}\in\mathcal{X}}Q_1(\pmb{x})\frac{\big[1 -\sum_{\mathbf{x}''\in\mathcal{X}}A(\mathbf{x}'',\mathbf{x})Q(\mathbf{x}''|\mathbf{x})\big]}{Q_1(\pmb{x})}\bigg\{\eta\big[1 -\sum_{\mathbf{x}''\in\mathcal{X}}A(\mathbf{x}'',\mathbf{x})Q(\mathbf{x}''|\mathbf{x})\big] -\tilde{p}(\pmb{x})\log\eta -\tilde{p}(\pmb{x})\bigg\} + \frac{\log\Gamma}{\Gamma} \nonumber \\    &\leq\frac{1}{\Gamma}\sum_{\pmb{x}\in\mathcal{X}}\sum_{\pmb{x}'\in N(\pmb{x})}
    Q_1(\pmb{x})Q_2(\pmb{x}')\frac{
    \tilde{p}(\pmb{x})Q(\mathbf{x}'|\mathbf{x})}{Q_1(\pmb{x})Q_2(\pmb{x}')}A(\mathbf{x}',\mathbf{x})\log\frac{A(\mathbf{x}',\mathbf{x})Q(\mathbf{x}'|\mathbf{x})}{\tilde{p}(\pmb{x}')} + \nonumber\\
    &\qquad +
    \frac{1}{\Gamma}\sum_{\pmb{x}\in\mathcal{X}}Q_1(\pmb{x})\frac{\big[1 -A(\mathbf{x}^*,\mathbf{x})Q(\mathbf{x}^*|\mathbf{x})\big]}{Q_1(\pmb{x})}\bigg\{\eta\big[1 -A(\mathbf{x}^*,\mathbf{x})Q(\mathbf{x}^*|\mathbf{x})\big] -\tilde{p}(\pmb{x})\log\eta -\tilde{p}(\pmb{x})\bigg\} + \frac{\log\Gamma}{\Gamma}    \end{align}
and by defining $\mathcal{M}(\pmb{x})=1-A(\mathbf{x}^*,\mathbf{x})Q(\mathbf{x}^*|\mathbf{x})$, we obtain case (b).
\end{proof}

{\section{Hyperparameters Used in the Experiments}
	\label{sec:D}}
\begin{itemize}
\item Learning rate $\eta=1e-2$ for SGD optimizer with momentum.
\item Burn-in iterations $K=2000$ (for Ising),  $K=24000$ (for RBM), $K=500$ (for UAI).
\item Iterations for sampling $30000$ (for Ising), $120000$ (for RBM), $10000$ (for UAI).
\item Batch size $N=30$ (for Ising), $N=16$ (for RBM), $N=5$ (for UAI).
\item MLP network with one hidden layer of $10$ neurons (for ISING and RBM) and monotonic network with 20 blocks of 20 neurons (for UAI). Refer to Appendix~\ref{sec:F} for additional experiments on UAI with MLP.
\end{itemize}

For experiments on RBM, we reused the code provided by~\cite{grathwohl2021oops} .

\begin{figure}[b]
     \centering
     \begin{subfigure}[b]{0.48\textwidth}
         \centering
         \includegraphics[width=0.9\textwidth]{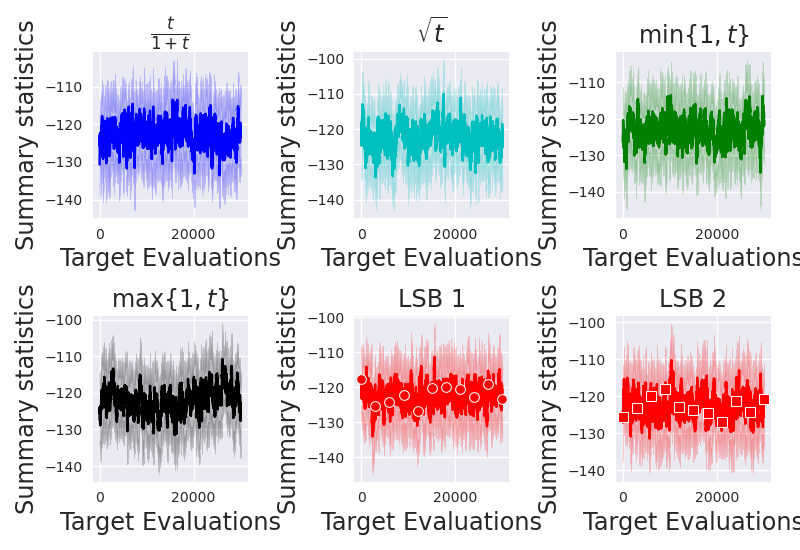}
         \caption{Case 1}
     \end{subfigure}%
     \begin{subfigure}[b]{0.48\textwidth}
         \centering
         \includegraphics[width=0.9\textwidth]{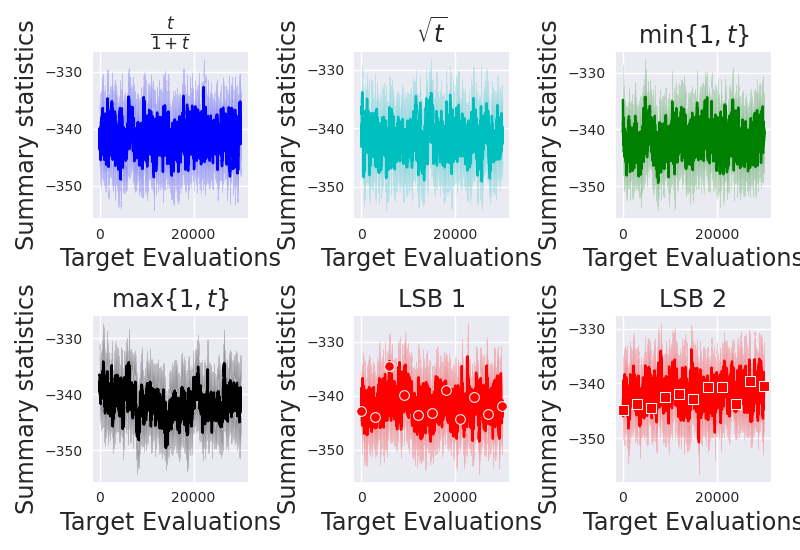}
         \caption{Case 2}
     \end{subfigure}\\
     \begin{subfigure}[b]{0.48\textwidth}
         \centering
         \includegraphics[width=0.9\textwidth]{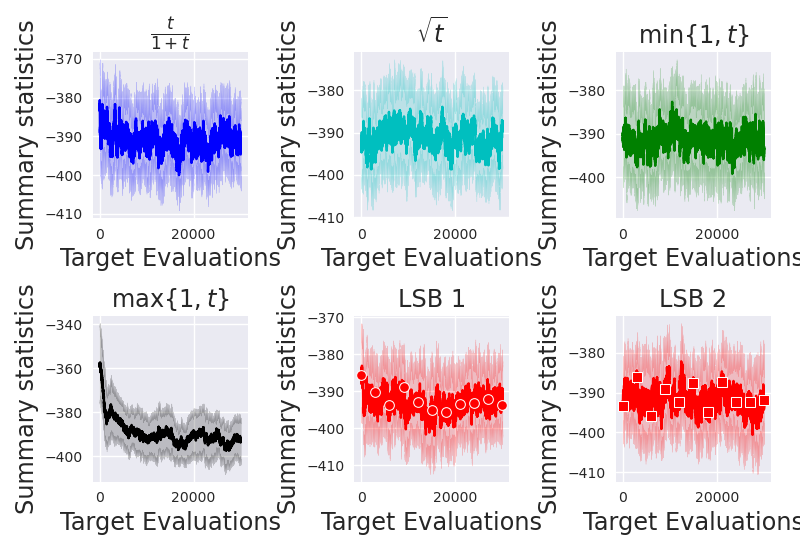}
         \caption{Case 3}
     \end{subfigure}%
     \begin{subfigure}[b]{0.48\textwidth}
         \centering
         \includegraphics[width=0.9\textwidth]{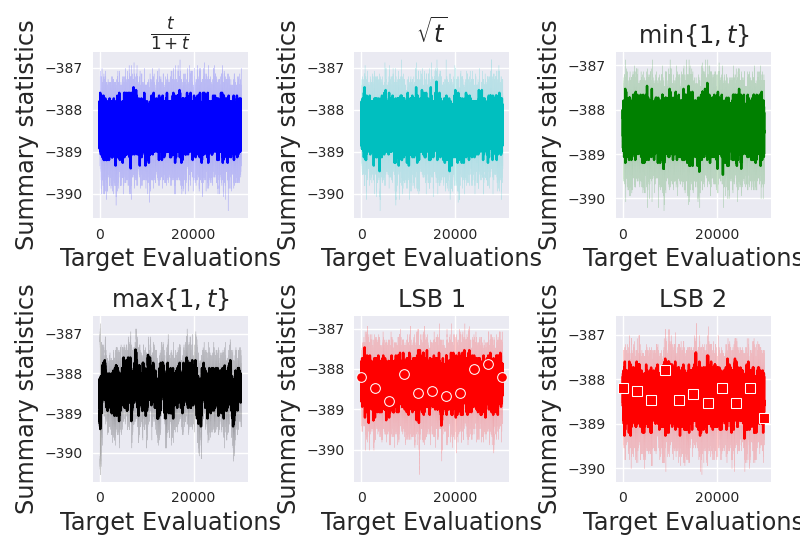}
         \caption{Case 4}
     \end{subfigure}%
     \caption{Traceplots on four cases of the Ising model ($30\times 30$) for the mixing phase. (a) Case 1: Independent-noisy, (b) case 2: Independent-clean, (c) case 3: Dependent-noisy, (d) case 4: Dependent-clean}
     \label{fig:trace}
\end{figure}

{\section{Further Results for Ising}\label{sec:E}}
See Figure~\ref{fig:trace}.
%
%
%
\begin{figure}
     \centering
     \begin{subfigure}[b]{0.48\textwidth}
         \centering
         \includegraphics[width=0.9\textwidth]{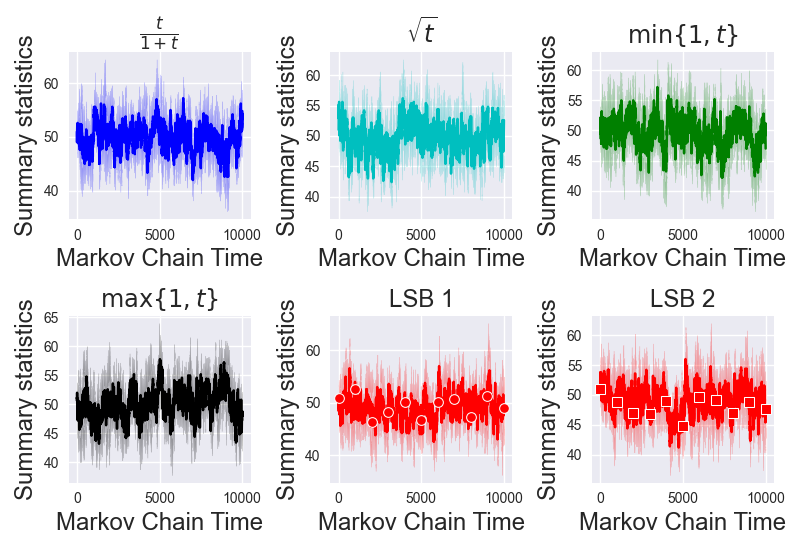}
         \caption{MN 1}
     \end{subfigure}%
     \begin{subfigure}[b]{0.48\textwidth}
         \centering
         \includegraphics[width=0.9\textwidth]{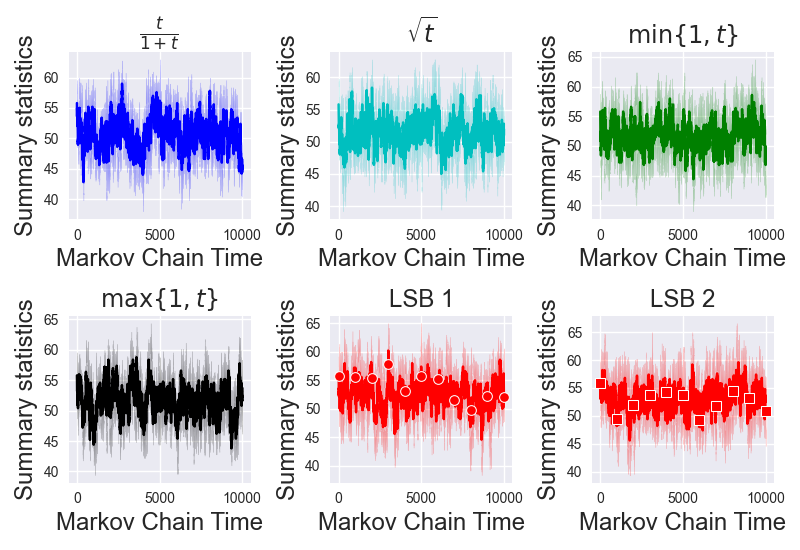}
         \caption{MN 8}
         \end{subfigure}%
    \caption{Traceplots on UAI benchmarks model ($100$ vars near-deterministic dependencies) for the mixing phase.}
     \label{fig:uai_trace}
\end{figure}

\begin{figure}[t]
     \begin{subfigure}[b]{0.33\linewidth}
         \centering
         \includegraphics[width=0.9\linewidth]{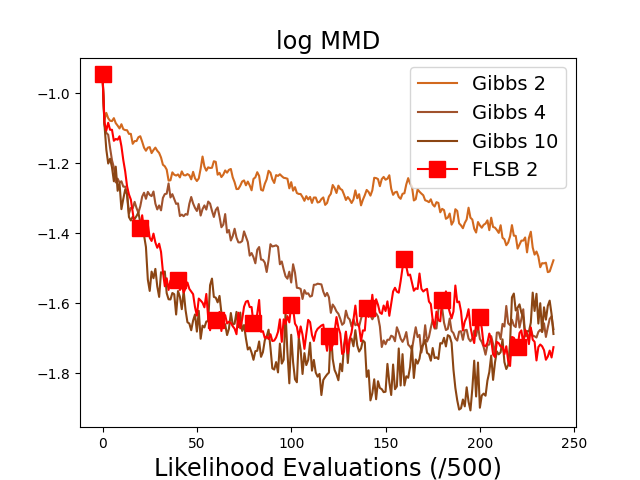}
     \end{subfigure} %
     \begin{subfigure}[b]{0.33\linewidth}
         \centering
         \includegraphics[width=0.9\linewidth]{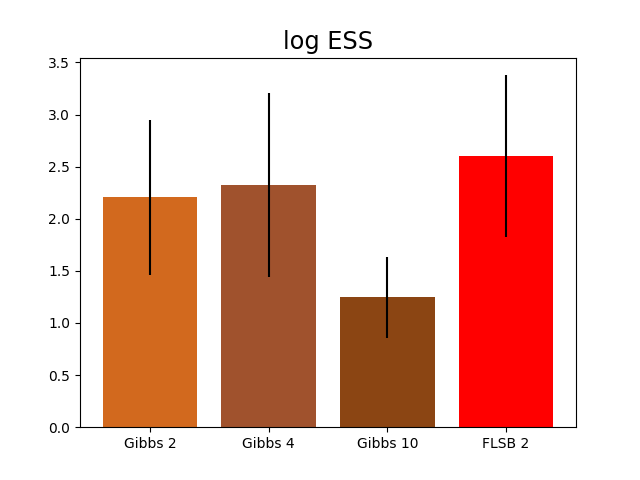}
     \end{subfigure}%
     \begin{subfigure}[b]{0.33\linewidth}
         \centering
         \includegraphics[width=0.9\linewidth]{img/RBM/COMP1_RBM_ess.png}
     \end{subfigure}%
     \caption{Comparison of FLSB 2 against block Gibbs sampling with block size of 2, 4 and 10 variables on RBMs.}
     \label{fig:rbm_gibbs}
\end{figure}

\begin{figure}[t]
     \begin{subfigure}[b]{0.33\linewidth}
         \centering
         \includegraphics[width=0.9\linewidth]{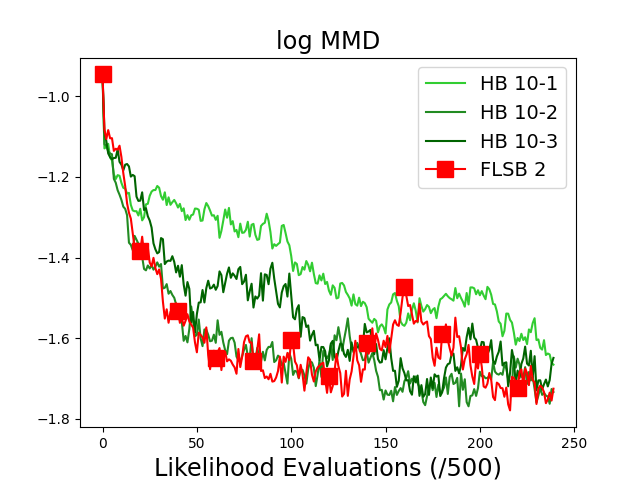}
     \end{subfigure} %
     \begin{subfigure}[b]{0.33\linewidth}
         \centering
         \includegraphics[width=0.9\linewidth]{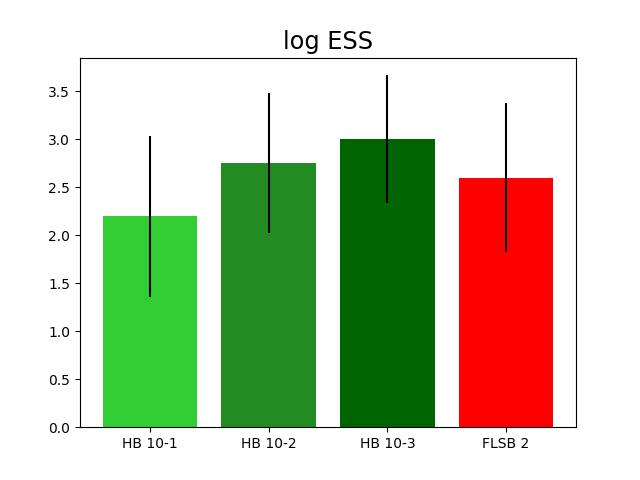}
     \end{subfigure}%
     \begin{subfigure}[b]{0.33\linewidth}
         \centering
         \includegraphics[width=0.9\linewidth]{img/RBM/COMP2_RBM_ess.png}
     \end{subfigure}%
     \caption{Comparison of FLSB 2 against the Hamming Ball sampler using 10 variables per block and updating 1, 2 and 3 variables per step.}
     \label{fig:rbm_hb}
\end{figure}

{\section{Additional Experiments for RBM}\label{sec:G}}
We provide additional comparisons of FLSB 2 against sampling strategies modifying more than 1 variable per step. In particular, we compare FLSB 2 against Gibbs 2, Gibbs 4 and Gibbs 10 and also compare FLSB 2 against HB-10-1, HB-10-2 and HB-10-3. Results are shown in Figures~\ref{fig:rbm_gibbs} and~\ref{fig:rbm_hb}, respectively. It is important to mention also that FLSB 2 uses a much smaller number of target likelihood evaluations compared to these previous approaches, as summarized in Table~\ref{tab:rbm}.
\begin{table*}
 \caption{Summary of the properties of different approaches.}
 \label{tab:rbm}
 \centering
\begin{small}
  \begin{tabular}{lcc}
    \toprule
    Method & Target likelihood evaluations per sampling step & Number of variables modified per sampling step \\
    \midrule
    Gibbs 2 & 4 & 2 \\
    Gibbs 4 & 16 & 4 \\
    Gibbs 10 & 1024 & 10 \\
    HB-10-1 & 20 & 1 \\
    HB-10-2 & 180 & 2 \\
    HB-10-3 & 960 & 3 \\
    FLSB 2 & \textbf{1} & 1 \\
\bottomrule
 \end{tabular}
\end{small}
\end{table*}

\begin{figure*}
     \centering
     \begin{subfigure}[b]{0.08\textwidth}
         \centering
         \includegraphics[width=0.9\textwidth]{img/LEGEND.png}
     \end{subfigure}%
     \begin{subfigure}[b]{0.22\textwidth}
         \centering
         \includegraphics[width=0.9\textwidth]{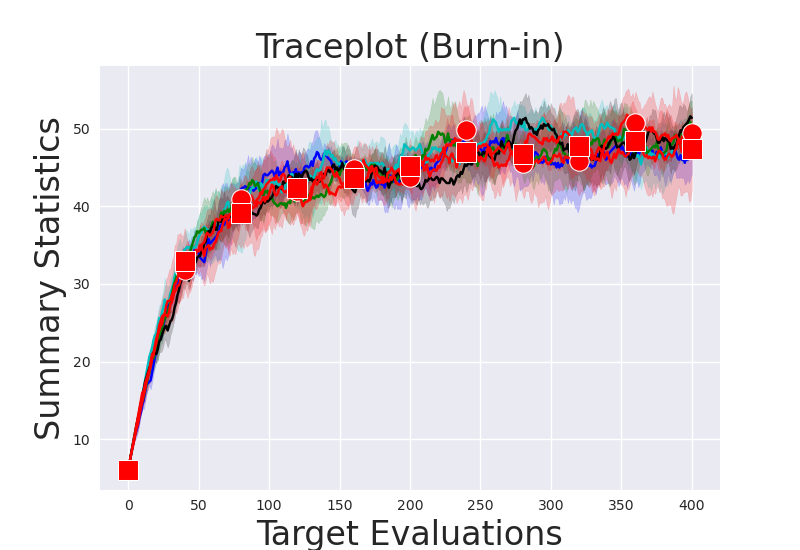}
         \caption{MN 1 (Burn-in)}
     \end{subfigure}%
     \begin{subfigure}[b]{0.22\textwidth}
         \centering
         \includegraphics[width=0.9\textwidth]{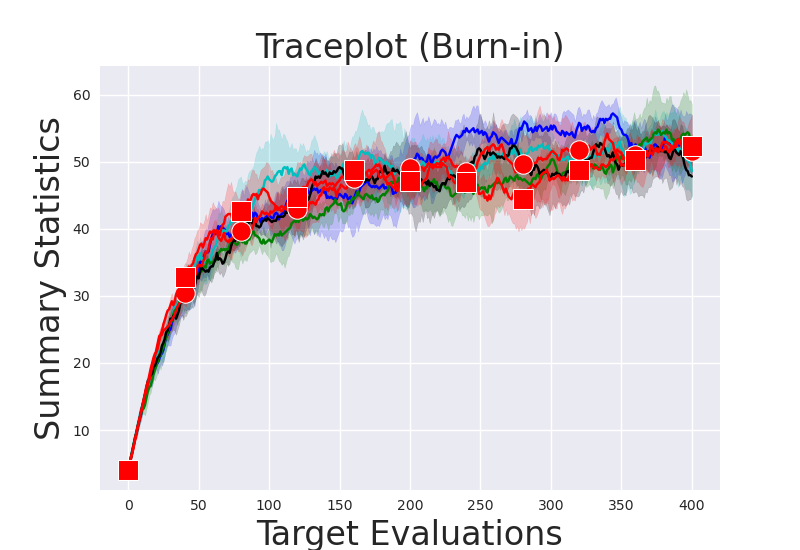}
         \caption{MN 2 (Burn-in)}
     \end{subfigure}%
     \begin{subfigure}[b]{0.22\textwidth}
         \centering
         \includegraphics[width=0.9\textwidth]{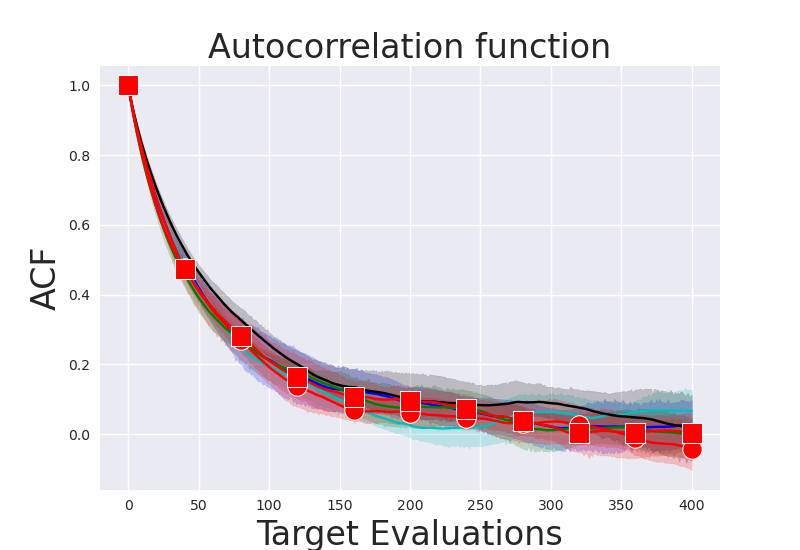}
         \caption{MN 1 (mixing)}
     \end{subfigure}%
     \begin{subfigure}[b]{0.22\textwidth}
         \centering
         \includegraphics[width=0.9\textwidth]{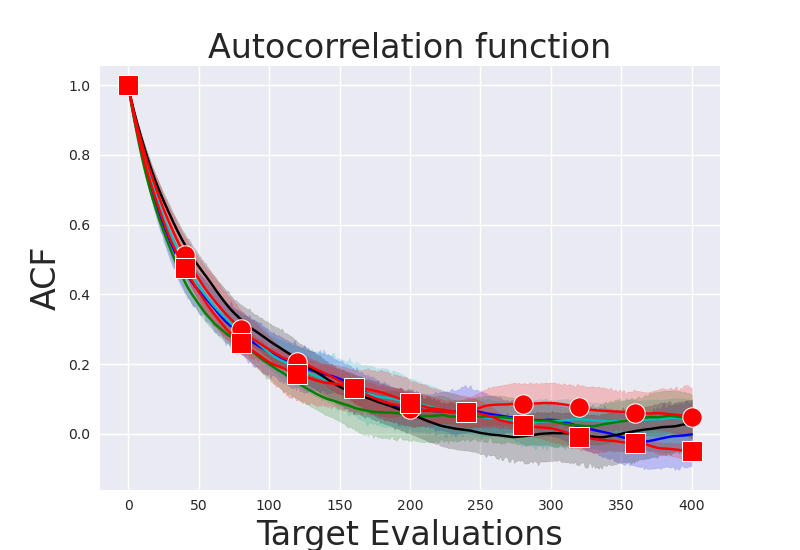}
         \caption{MN 2 (mixing)}
     \end{subfigure}%
     \caption{Samplers' performance on Markov networks from UAI competition ($100$ variables) with MLP network. (a)-(b) are the traceplots for the burn-in phase, while (c)-(d) are the autocorrelation function for the mixing one}
     \label{fig:uai_mlp}
\end{figure*}

\begin{figure}
     \centering
     \begin{subfigure}[b]{0.24\textwidth}
         \centering
         \includegraphics[width=0.9\textwidth]{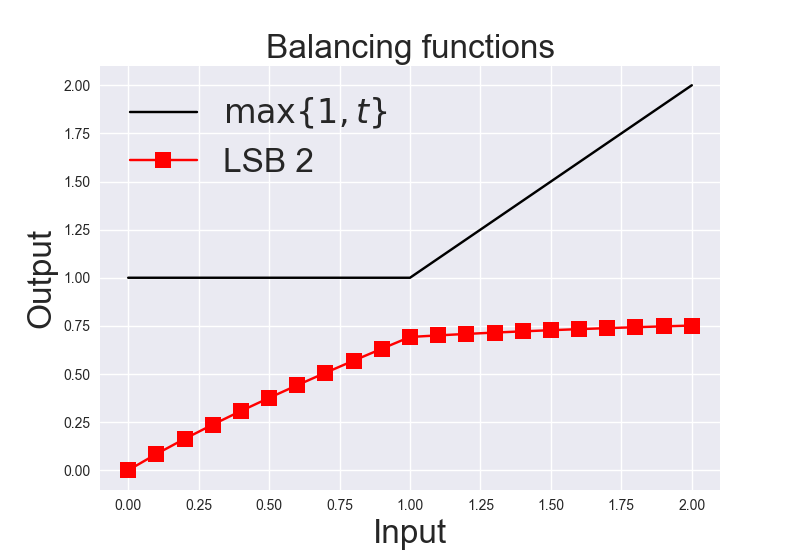}
         \caption{Iteration 0}
     \end{subfigure}%
     \begin{subfigure}[b]{0.24\textwidth}
         \centering
         \includegraphics[width=0.9\textwidth]{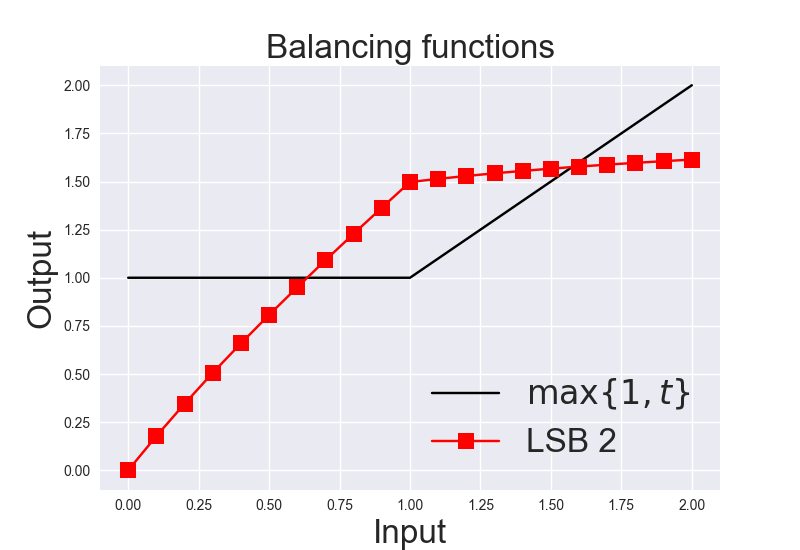}
         \caption{Iteration 1000}
     \end{subfigure}
     \begin{subfigure}[b]{0.24\textwidth}
         \centering
         \includegraphics[width=0.9\textwidth]{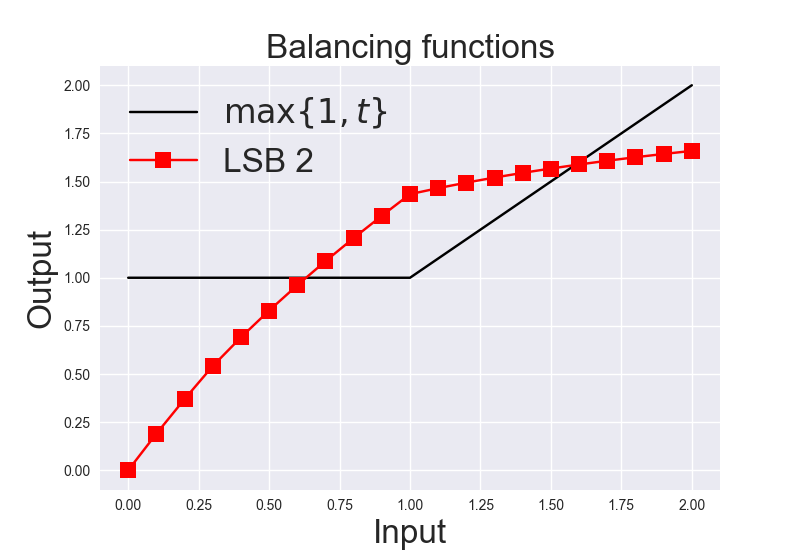}
         \caption{Iteration 2000}
     \end{subfigure}%
     \begin{subfigure}[b]{0.24\textwidth}
         \centering
         \includegraphics[width=0.9\textwidth]{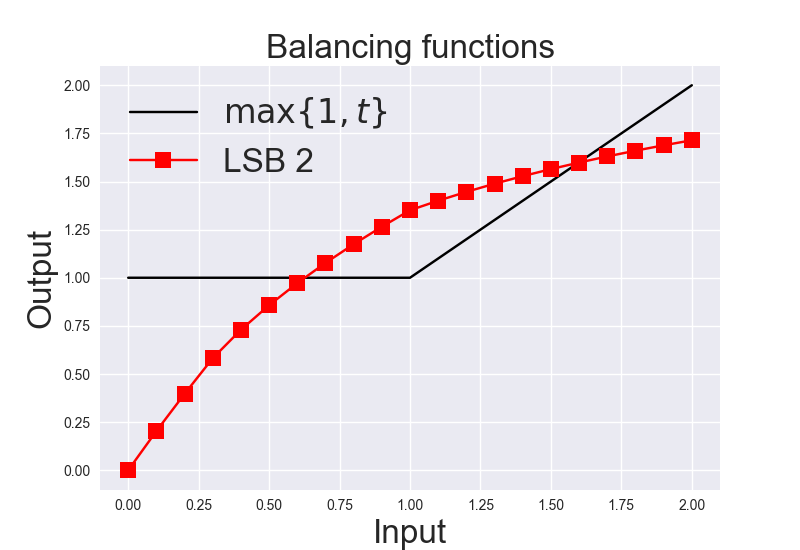}
         \caption{Iteration 3000}
     \end{subfigure}\\
     \begin{subfigure}[b]{0.24\textwidth}
         \centering
         \includegraphics[width=0.9\textwidth]{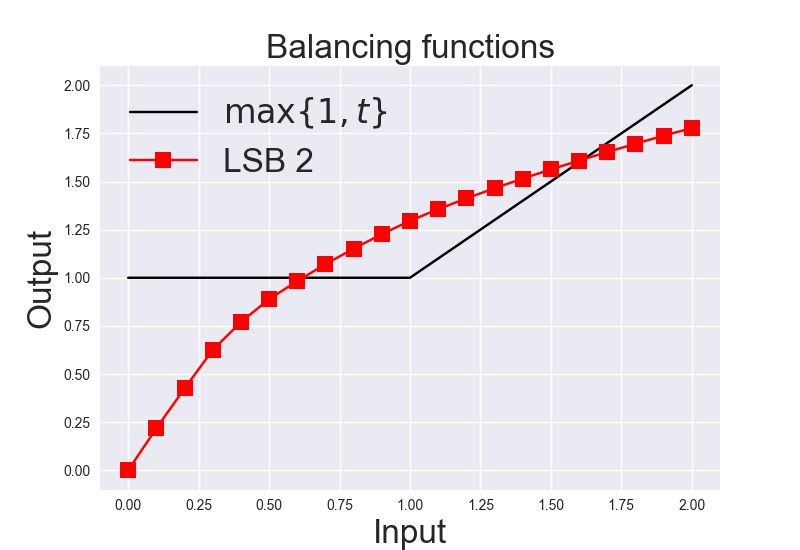}
         \caption{Iteration 4000}
     \end{subfigure}%
     \begin{subfigure}[b]{0.24\textwidth}
         \centering
         \includegraphics[width=0.9\textwidth]{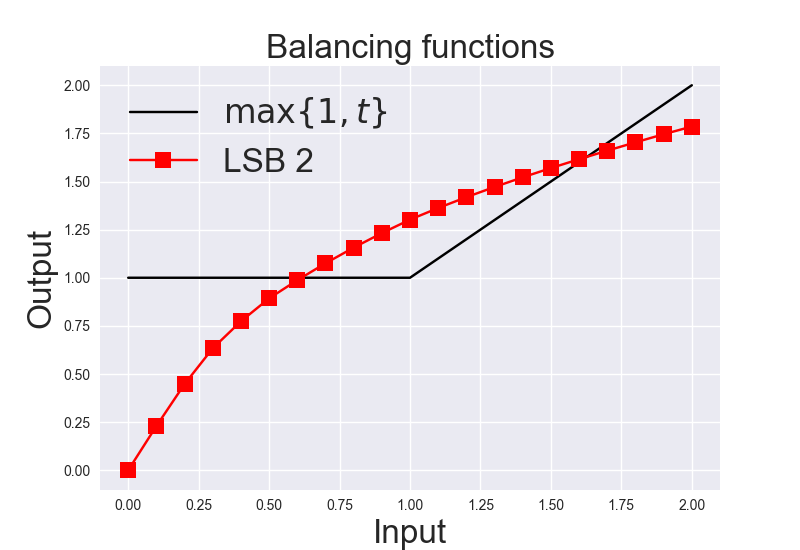}
         \caption{Iteration 5000}
     \end{subfigure}
     \begin{subfigure}[b]{0.24\textwidth}
         \centering
         \includegraphics[width=0.9\textwidth]{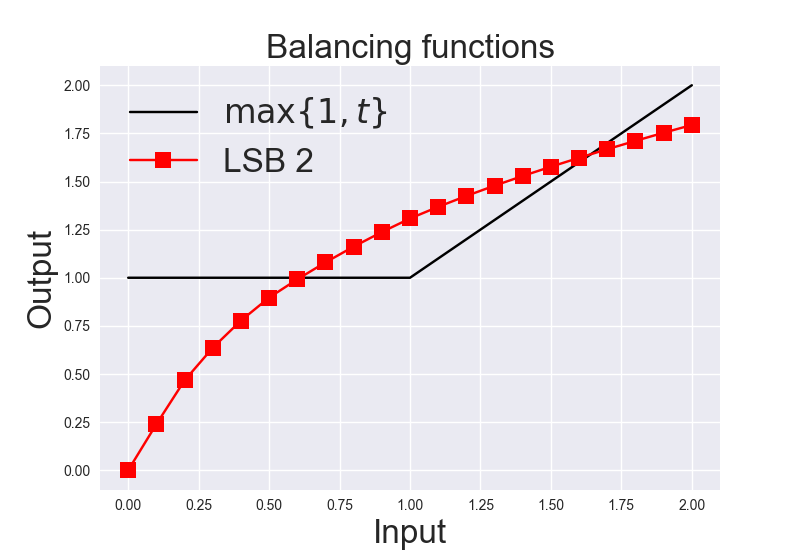}
         \caption{Iteration 6000}
     \end{subfigure}%
     \begin{subfigure}[b]{0.24\textwidth}
         \centering
         \includegraphics[width=0.9\textwidth]{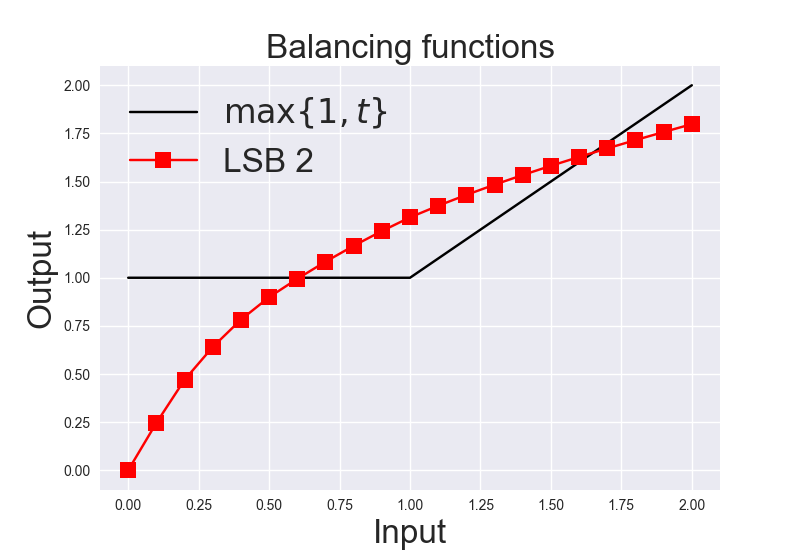}
     \end{subfigure}
     \caption{Training the monotonic network to match $\max\{1,t\}$ balancing function.}
     \label{fig:init}
\end{figure}

{\section{Additional Experiments on UAI}\label{sec:F}}
We repeated the experiments on UAI using a MLP network with one hidden layer of $10$ neurons, shown in Figure~\ref{fig:uai_mlp}. We found that in this domain characterized by large close-to-zero mass regions, the initial random input $\mathbf{x}$ is likely to fall outside the distribution support. In such case, there is no need to have balancing functions with $g(0)>0$. Notably, this doesn't occur when using a monotonic network, as inspected by the experiments in Figure~\ref{fig:init}, where we train the monotonic network to match a function like $\max\{1,t\}$.

%
%

%
%

\end{document}